\documentclass[10pt,twocolumn,letterpaper]{article}

\usepackage[pagenumbers]{cvpr} 

\usepackage{graphicx}
\usepackage{amsmath}
\usepackage{amssymb}
\usepackage{booktabs}
\usepackage{cuted}
\usepackage{multirow} 
\usepackage{diagbox} 
\usepackage{enumitem}
\usepackage{color}
\usepackage{cite}
\usepackage[table,xcdraw]{xcolor}
\usepackage{colortbl}
\usepackage{booktabs}
\usepackage{multirow} 
\usepackage{rotating}
\usepackage{caption}
\usepackage{subcaption}

\usepackage{amsmath,amsfonts,bm}

\def\eqref#1{equation~\ref{#1}}

\def\1{\bm{1}}

\def\vp{{\bm{p}}}

\def\vx{{\bm{x}}}
\def\vy{{\bm{y}}}

\DeclareMathAlphabet{\mathsfit}{\encodingdefault}{\sfdefault}{m}{sl}
\SetMathAlphabet{\mathsfit}{bold}{\encodingdefault}{\sfdefault}{bx}{n}

\def\gD{{\mathcal{D}}}

\def\gY{{\mathcal{Y}}}

\newcommand{\E}{\mathbb{E}}

\newcommand{\softmax}{\mathrm{softmax}}

\newcommand{\ti}{\text{i}}
\newcommand{\norm}[1]{\left\lVert#1\right\rVert}

\newtheorem{definition}{Definition}

\usepackage[pagebackref,breaklinks,colorlinks]{hyperref}

\usepackage[capitalize]{cleveref}
\crefname{section}{Sec.}{Secs.}
\Crefname{section}{Section}{Sections}
\Crefname{table}{Table}{Tables}
\crefname{table}{Tab.}{Tabs.}

\def\cvprPaperID{*****} 
\def\confName{CVPR}
\def\confYear{2022}

\begin{document}

\title{Smoothing Matters: Momentum Transformer\\for Domain Adaptive Semantic Segmentation}

\author{
 Runfa Chen$^1$\thanks{Email: crf21@mails.tsinghua.edu.cn}, Yu Rong$^3$, Shangmin Guo$^4$, Jiaqi Han$^1$, Fuchun Sun$^1$\thanks{Corresponding author: Fuchun Sun (fcsun@mail.tsinghua.edu.cn).}, Tingyang Xu$^3$, Wenbing Huang$^2$\\
 $^1$Institute for Artificial Intelligence (THUAI),\\
 Beijing National Research Center for Information Science and Technology (BNRist),\\
 State Key Lab on Intelligent Technology and Systems,\\ Department of Computer Science and Technology, Tsinghua University\\ 
 $^2$Institute for AI Industry Research (AIR), Tsinghua University\\
 $^3$Tencent AI lab\\
 $^4$School of Informatics, University of Edinburgh\\
 }

\maketitle

\begin{strip}
    \centering
    \vspace{-40pt}
    \includegraphics[trim={0cm 0cm 0cm 0cm},clip,width=\textwidth]{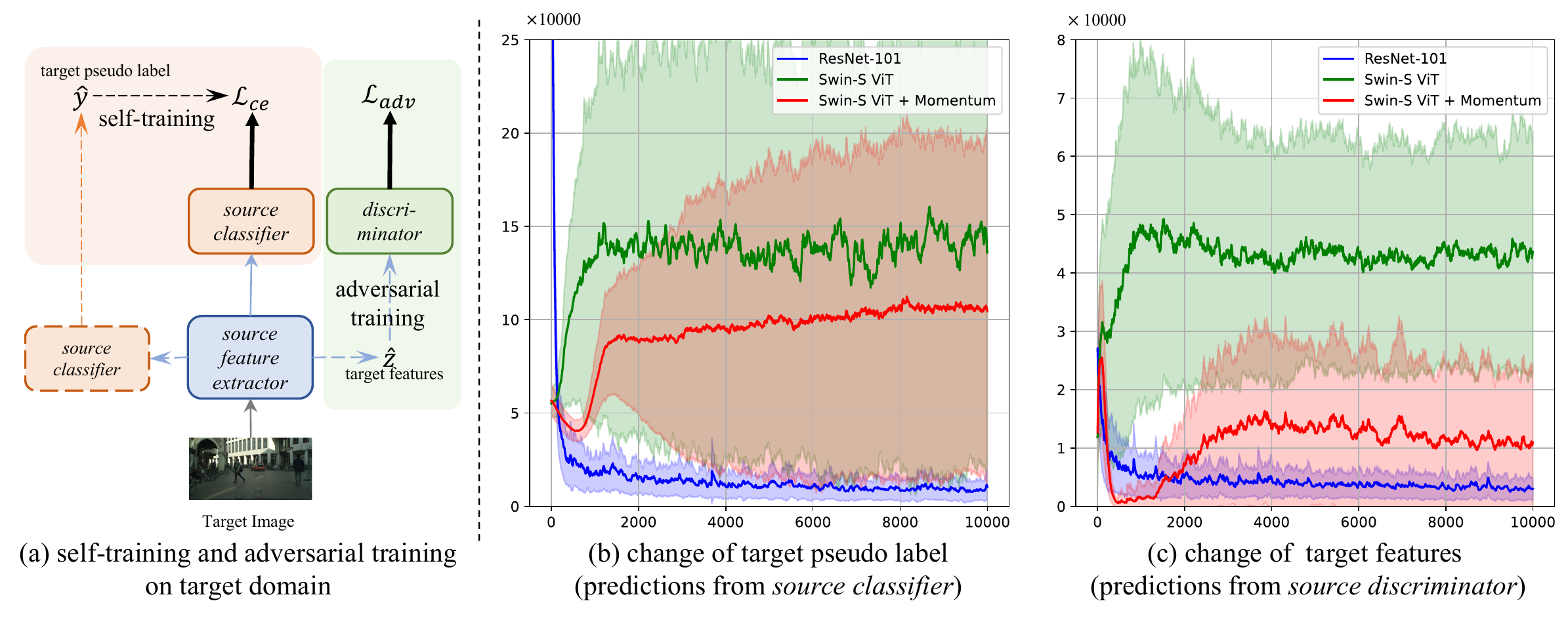}
    \vspace{-2.em}
    \captionof{figure}{The conventional training paradigm and change of predictions over training iterations. 
     We sampled images from target domain, Cityscapes~\cite{cordts2016cityscapes}, and plot the L1-distance between the predictions over images on iteration $t$ and next iteration $t+1$ as the lines. 
    The standard deviation of the distances are indicated by the corresponding shadow areas around the lines.  
    See \textsection~\ref{sec:Momentum} for details.
    In both (b) and (c) diagrams, higher value means more unsmooth learning dynamics, thus more high-frequency components, since the model always significantly change its prediction iteration by iteration.
    }
    \label{fig:title}
\end{strip}

\begin{abstract}
After the great success of Vision Transformer variants (ViTs) in computer vision, it has also demonstrated great potentials in domain adaptive semantic segmentation. Unfortunately, straightforwardly applying local ViTs in domain adaptive semantic segmentation does not bring in expected improvement. We find that the pitfall of local ViTs is due to the severe \emph{high-frequency components} generated during both the pseudo-label construction and features alignment for target domains. These high-frequency components make the training of local ViTs very unsmooth and hurts their transferability. In this paper, we introduce a low-pass filtering mechanism, momentum network, to smooth the learning dynamics of target domain features and pseudo labels. Furthermore, we propose a dynamic of discrepancy measurement to align the distributions in the source and target domains via dynamic weights to evaluate the importance of the samples. After tackling above issues, extensive experiments on sim2real benchmarks show that the proposed method outperforms the state-of-the-art methods. Our codes are available at \href{https://github.com/alpc91/TransDA}{https://github.com/alpc91/TransDA}.
\end{abstract}


\section{Introduction}
\label{sec:intro}

Semantic segmentation~\cite{xiao2018unified,long2015fully,zhao2017pyramid,chen2017deeplab}, as one of the most central tasks in computer vision, aims to label the semantic content of an image pixel by pixel. 
It requires dense annotations of training images and becomes more challenging when it is difficult to obtain these dense labels in practice.
A commonly recognised solution is the domain adaptive semantic segmentation~\cite{richter2016playing,ros2016synthia}, which usually first trains models on abundantly labeled images simulated by virtual environment, then carries out the simulation-to-reality (sim2real) domain adaptation (DA) to adapt the model on real-world images.

Given the past success of Convolutional Neural Networks (CNNs) on many computer vision (CV) tasks~\cite{he2016deep, long2015fully, ren2015faster}, plenty of works~\cite{hoffman2016fcns,chen2017no,tsai2018learning,luo2019taking,hoffman2018cycada,wang2020classes,yang2020fda,zheng2021rectifying,zhang2021prototypical, guo2021metacorrection,araslanov2021self,ma2021coarse}  resort to CNNs as the semantic segmentation function $f_{\theta}$. 
Although these conventional CNN-based backbones obtained decent performance on various benchmarks, recent works (\eg DRT~\cite{li2021dynamic}) show that they are still unable to tackle domain conflicts, and suffering from performance bottleneck.
To make further improvement, a natural alternative backbone is the Transformer model proposed by \cite{vaswani2017attention} which has shown to be more powerful than CNNs in various areas: i) natural language processing (NLP) tasks, \eg~\cite{brown2020language,kenton2019bert}; ii) CV tasks, \eg~\cite{dosovitskiy2020image,yuan2021tokens,wang2021pyramid,touvron2021training,liu2021Swin}.
More importantly, existing works such as~\cite{liu2021Swin,xie2021segformer,guo2021sotr} have also shown the great potential of the Vision Transformer variants (ViTs) for semantic segmentation.

Inspired by the above success of ViTs, we propose a domain adaptive semantic segmentation framework based on local ViTs without incurring substantial training efforts in this work. We employ local ViTs as backbone and adapt a simple semi-supervised approach, self-training~\cite{chen2011co,lee2013pseudo} along with adversarial training~\cite{hoffman2016fcns,chen2017no,tsai2018learning,wang2020classes} widely-used in DA (Figure~\ref{fig:title}(a)).
Since there is no ground-truth for target domain, in self-training, target pseudo labels are crucial for the classifier to make dense prediction, and in adversarial training, target features are essential for the discriminator to align features distribution.
Based on the above basic framework, our contribution is two-fold as follows.

As far as we know, this is the first work that leverages local ViTs as backbone and explores its potential on domain adaptive semantic segmentation. 
Unfortunately, through our experiments, we find that direct self-training along with adversarial training upon ViT segmentation network does not bring in expected improvement, as shown in Table~\ref{tab:Component}. 
Our first contribution is the discovery of \emph{the high-frequency component} problem of local ViTs in pseudo-label generation and feature alignment for target domain.
The statistics about the change of predictions over training iterations are illustrated in Figure\ref{fig:title}.
By comparing the change of predictions from different backbones, we can see that the predictions on target domain of local ViTs (Swin-S in our specific case) vibrate more drastically than CNNs (ResNet-101 in our specific case), as indicated by the larger changes of predictions over iterations.

Our second contribution is a solution to tackle the above issues.
We formalize the self-training and adversarial training as knowledge distillation~\cite{hinton2015distilling} and features alignment respectively, \emph{i.e.} the network from source domain is the teacher network to provide pseudo labels and features for target domain.
Following the findings from \cite{ren2021better} that low-pass filtering the predictions from teacher can improve the quality of supervision, we also introduce a low-pass filtering into features and pseudo labels for target domain.
However, their original implementation maintains a look-up table which stores predictions for all training example, which is infeasible in semantic segmentation as it requires huge storage overhead for all pixels.
Therefore, we propose an indirect low-pass filter, the momentum network, to smooth the learning dynamics for target domain. 
Specifically, the feature extractor and classifier for target domain is updated with momentum copy from source domain, as illustrated in Figure~\ref{fig:Mo}. Notably, jointly smoothing the pseudo labels and features is essential. After applying momentum network, the change of predictions over training iterations is indeed significantly suppressed in Figure~\ref{fig:title}.

To further smooth the learning dynamics and suppress noise, we propose a dynamic adversarial training strategy, dynamic of discrepancy measurement.
In particular, the alignment is guided by a score function $w(\vx)$ that adaptively balances the importance of each sample during the adaptation process.
Through experiments, we find that this strategy can further improve the performance of ViTs-based backbone, whereas it doesn't work with CNN-based backbones.
We argue that the high-frequency component problem is specific for ViTs but not CNNs, which is supported by both the smaller changes of predictions from CNNs illustrated in Figure~\ref{fig:title}(b)(c) and the different feature representations learnt by CNNs as illustrated in Figure~\ref{fig:umap}.

In summary, we propose a novel framework based on local ViTs dubbed TransDA, via a simple yet effective training paradigm. 
Exploring this new baseline shows that it's an urge to develop new methods based on a new backbone.
After solving the high-frequency component issues during adaptation, 
our experiments on two standard sim2real benchmarks, GTA5\cite{richter2016playing}$\rightarrow$Cityscapes\cite{cordts2016cityscapes} and SYNTHIA\cite{ros2016synthia}$\rightarrow$Cityscapes, demonstrate that TransDA achieves attractive and robust performance in terms of best and averaged mIoU.

\section{Related Work}

\paragraph{Domain adaptation.}
Domain adaptation deals with the scenario where a labeled source domain and an unlabeled target domain are provided, and certain shift in distribution exists in between~\cite{shimodaira2000improving,PFAN_2019_CVPR,hoffman2016fcns}. 
To bridge the gap, a line of work leverages the idea of adversarial learning~\cite{goodfellow2014generative}, \ie train the target feature extractor with a domain classifier under an adversarial schedule~\cite{ganin2016domain,tzeng2017adversarial,you2019universal}. 
UniDA~\cite{you2019universal} proposes a weighting mechanism to identify private classes. 
In contrast to the universal DA , we adopt this similar method to dynamically adjust the pixel-by-pixel adversarial weight for closed set DA and semantic segmentation.
DRT\cite{li2021dynamic} abandons conventional CNNs backbone and  present dynamic transfer to address domain conflicts, where the model parameters are adapted to sample.  It inspires us to explore a new backbone in domain adaptive semantic segmentation.

\vspace{-1.em}
\paragraph{Domain adaptive semantic segmentation.}
Large process has been made in semantic segmentation with the paradigm of domain adaptation. 
Many works~\cite{hoffman2016fcns, tsai2018learning,luo2019taking,hoffman2018cycada,wang2020classes,chen2017no} utilize standard adversarial learning to enhance the performance. 
Among them, AdaptSeg~\cite{tsai2018learning}, which is the first work to replace VGG with ResNet backbone in this task.
Besieds, \cite{sun2019not} assigns sample-wise weight scores for cross-entropy loss to suppress the negative effect in the transfer process. 
In contrast, our dynamic discrepancy is designed for adversarial loss.
Besides, self-training is another technique, which generates pseudo labels for making use of the unlabeled data. 
For instance, Seg-Uncertainty~\cite{zheng2021rectifying} predicts the confidence for the pseudo labels; 
ProDA~\cite{zhang2021prototypical} relies on prototypes to further correct the pseudo labels during training in an online manner; 
MetaCorrection~\cite{guo2021metacorrection} depicts the noise distribution of pseudo labels via meta-learning; 
SAC~\cite{araslanov2021self} proposes an augmentation consistency approach trained on co-evolving pseudo labels. Akin to the momentum network in SAC, our momentum network requires smooth pseudo labels and features for target domain at the same time, without incurring substantial effort, \eg, thresholds and focal loss, importance sampling, etc.
Self-training is widely used in conjunction with adversarial training in recent works\cite{choi2019self,kim2020learning,mei2020instance,wang2020classes,yang2020fda,zheng2021rectifying}. In this paper, we successfully explore the potential of this kind of joint training paradigm for local ViTs in domain adaptation.

\vspace{-1.em}
\paragraph{Other related works.} Recently, Transformer\cite{vaswani2017attention} and dynamic networks~\cite{han2021dynamic} become more popular in deep learning.
Deeper discussion about the relationship between Transformer and dynamic networks can be found in \cite{han2021dynamic,han2021demystifying}.
Pioneered by ViT~\cite{dosovitskiy2020image}, many Transformer-based vision backbones~\cite{yuan2021tokens,wang2021pyramid,touvron2021training,liu2021Swin,xie2021segformer,guo2021sotr} have been proposed and achieved superior performance over a wide range of CV tasks.
Particularly, by introducing shifted windows which capture local information to the self-attention module, Swin-Transformer~\cite{liu2021Swin} became a representative of the local ViTs~\cite{chu2021twins,vaswani2021scaling,han2021demystifying}. 
Despite the prevalence of Transformers in CV tasks, there is still limited attempt of them in domain adaptation.

\section{TransDA} 
In this section, we first illustrate the overall architecture of our method, then provide more details about each component, \ie, local ViT segmentation network, momentum network, and dynamic of discrepancy measurement, in the corresponding subsections.

\begin{figure*}[t]
    \centering
    \includegraphics[width=1.0\linewidth]{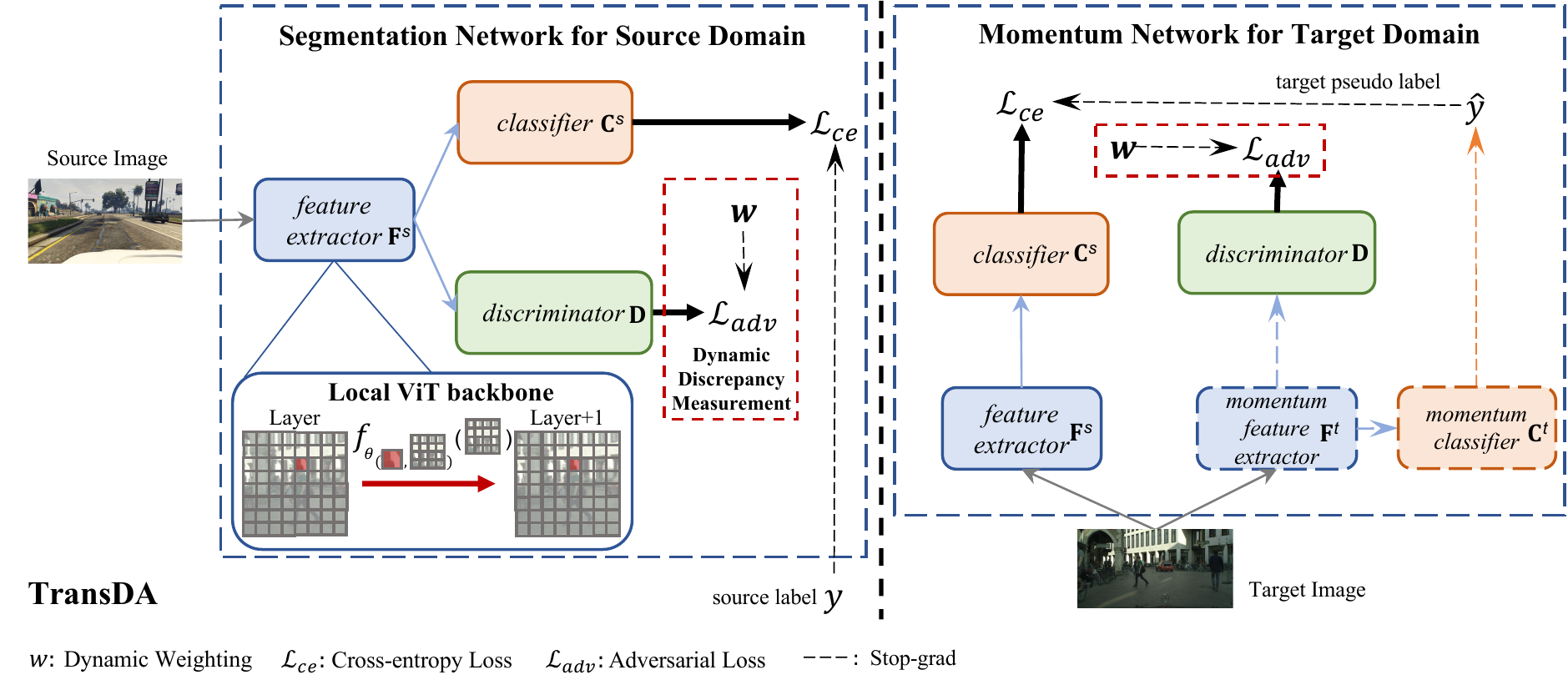}
    \vspace{-2em}
    \caption{ Momentum Transformer Domain Adaptive Semantic Segmentation } 
    \label{fig:arch}
\end{figure*}

\subsection{General Formulation}
In domain adaptive semantic segmentation, we have a source domain $\gD^s=\{\vx_i^s, \vy_i^s\}_{i=1}^{N^s}$ that consists of $N^s$ labeled images and a target domain  $\gD^t=\{\vx_i^t\}_{i=1}^{N^t}$ of $N^t$ unlabeled images. 
Let $\gY^s$ and $\gY^t$ denote the label spaces of the source and target spaces, respectively.   For simplicity, the superscripts $s$ and $t$ denote the source and target domains, respectively.
We use $\gY^c = \gY^s \cap \gY^t$ to denote the common label set shared by both domains, and $\gY^{sp} = \gY^s \backslash \gY^c$ and $\gY^{tp} = \gY^t \backslash \gY^c$ to represent the source and target private parts, respectively, following the setting of UniDA~\cite{you2019universal}. 
We denote $K^c$ the number of classes in $\gY^c$.
In what follows, we utilize the symbols $x(i,j)$ (resp. $y(i,j)$) to denote each pixel of the input image (resp. the output dense predictions) at grid $(i,j)$, and $x$ (resp. $y(x)$) to represent arbitrary pixel by omitting the pixel index.

Following \cite{han2021demystifying} and attention~\cite{vaswani2017attention} definition, 
$\theta(\cdot)$ varies with the input. 
From this, we can define the segmentation network using local ViTs as $f_{\theta(\cdot)}$.
Based on local ViTs, the goal of domain adaptive semantic segmentation is to learn a segmentation network $f_{\theta(\cdot)}^t$ with parameter function $\theta(\cdot)$ for target domain that can be generalized from source domain $f_{\theta(\cdot)}^s$. 
This implies 
We decompose it into two sub-goals: i) fitting the source-domain data, and ii) aligning the distributions of features between the source and target domains. 
The formal definition is given:
\begin{definition}{(Transformer Domain Adaptive Semantic Segmentation)} 
Given the denotations defined before, the segmentation network $f_{\theta(\cdot)}$ is learned by 
\begin{equation}
\label{eq.problem}
\footnotesize
\aligned
    \min_{\theta(\cdot)} \quad &\E_{\substack{\vx^s,\vy^s\sim\gD^s \\ \vx^t\sim\gD^t}}[\text{H}(f^s_{\theta(\vx^s;\vartheta^s)}(\vx^s), \vy^s) +\text{H}(f^s_{\theta(\vx^t;\vartheta^s)}(\vx^t), \hat{\vy}^t) \\
    &+ \text{Dis}(f^s_{\theta(\vx^s;\vartheta^s)}(\vx^s), f^t_{\theta(\vx^t;\vartheta^t)}(\vx^t))],\\
    \operatorname{ s.t. } \quad & \vartheta^t \leftarrow \vartheta^s,\\
      & \hat{\vy}^t \leftarrow f^t_{\theta(\vx^t;\vartheta^t)}(\vx^t),
\endaligned
\end{equation}
where, $\theta(\cdot)$ is a function of the input sample, $\E[\cdot]$ computes the expectation, $\text{H}(\cdot)$ can usually be implemented as the cross-entropy loss, and $\text{Dis}(\cdot)$ measures the discrepancy between domains. 
The \emph{pseudo labels} $\hat{\vy}^t$ is constructed from the most probable class predicted by $f_{\theta(\cdot)}^t$, which is directly copied from $f_{\theta(\cdot)}^s$ where  $\vartheta^s$ are the parameters of $f_{\theta(\cdot)}^s$ and $\vartheta^t$are those of $f_{\theta(\cdot)}^t$.  
\end{definition}

In practice,  $\text{H}(\cdot)$ and $\text{Dis}(\cdot)$ do not share the exact function $f_{\theta(\cdot)}$ in Eq.~\ref{eq.problem}; for example, as what we do in our implementation later, the outputs by $f_{\theta(\cdot)}$ are treated as the hidden features and will be stacked with different layers before the inputs for  $\text{H}(\cdot)$ and $\text{Dis}(\cdot)$. Here, for brevity, we omit this kind of nonessential difference, and use Eq.~\ref{eq.problem} in its current form. 

As discussed later in Sec~\ref{sec:Momentum}, to smooth the learning dynamics on target domain, we propose a momentum network to work with ViTs, \ie, Momentum Transformer. 
Moreover, the discrepancy loss $\text{Dis}(\cdot)$ penalizes all input pairs as a whole, which will compromise the discrimination between the samples of different classes in either the source or target domain. 
Therefore, we correlate both the segmentation network and the discrepancy measurement with the input samples.  
The formal formulation of the problem is given below.
\begin{definition}{(Momentum Transformer Domain Adaptive Semantic Segmentation)} 
Given the denotations defined before, the segmentation network $f_{\theta(\cdot)}$ and the weight $w$ are learned by 
\begin{equation}
\label{eq.new-problem}
\footnotesize
\aligned
    \min_{\theta(\cdot), \color{red}{w(\cdot,\cdot)}} &\E_{\substack{\vx^s,\vy^s\sim\gD^s \\ \vx^t\sim\gD^t}}[\text{H}(f^s_{\theta(\vx^s;\vartheta^s)}(\vx^s), \vy^s) +\text{H}(f^s_{\theta(\vx^t;\vartheta^s)}(\vx^t), \color{red}\hat{\vy}^t\color{black}) \\
    &+ \text{Dis}_{\color{red}{w(\color{black}\vx^s,\vx^t\color{red})}\color{black}}(f^s_{\theta(\vx^s;\vartheta^s)}(\vx^s), \color{red}f^t_{\theta(\color{black}\vx^t\color{red};\vartheta^t)}\color{black}(\vx^t))],\\
    \operatorname{ s.t. } \quad & \color{red}\vartheta^t \leftarrow m \vartheta^t +(1-m) \vartheta^s,\\
      & \color{red}\hat{\vy}^t \leftarrow f^t_{\theta(\color{black}\vx^t;\color{red}\vartheta^t)}\color{black}(\vx^t),
\endaligned
\end{equation}
Here, $\theta(\cdot)$ is a function of the input sample, and $\text{Dis}(\cdot)$ is \textbf{correlated with the weight $w(\cdot,\cdot)$}.  
The \emph{pseudo labels} $\hat{\vy}^t$ is constructed from the most probable class predicted by $f_{\theta(\cdot)}^t$, which is \textbf{updated with momentum copy} from $f_{\theta(\cdot)}^s$ where $m\in [0, 1)$ is a momentum coefficient. 
\end{definition}

In the following subsections, we will introduce the details of the local ViT segmentation network $f^s_{\theta(\cdot)}$, the momentum network  $f^t_{\theta(\cdot)}$ and the dynamic of discrepancy measurement $\text{Dis}_{w(\vx^s, \vx^t)}$.

\subsection{Local ViT Segmentation Network for Domain Adaptation}
\label{sec:dsn}
In this paper, our segmentation network $f_{\theta(\cdot)}$ via a typical Local ViT model -- Swin Transformer for Semantic Segmentaion~\cite{liu2021Swin}.
It includes a feature extractor $\mathbf{F}$ and a classifier $\mathbf{C}$ that will be used for the formulation of the loss $\text{H}(\cdot)$ in Eq.~\ref{eq.new-problem}.
The segmentation categorical prediction is then:
\begin{equation}
\aligned
    \vp^s &= [p^s_1,\dots,p^s_{K^c}] = \mathbf{C}^s(\mathbf{F}^s(\vx^s)) = f_{\theta(\vx^s)}^s(\vx^s)
\endaligned
\label{eq.ps}
\end{equation}
With the above function, the objective $\text{H}(\cdot)$ in Eq.~\ref{eq.new-problem} can be formally defined:
\begin{equation}
\aligned
    \mathcal{L}_{ce}^{source} &=  -\mathbb{E}_{\gD^s}[ \sum_{k=1}^{K^c} y_{k}^s \log \left(p^s_{k}\right)], 
\endaligned
\label{eq.ces}
\end{equation}
where $p^s_k$ and $y_k^s$ denote the predicted outputs and ground-truth, respectively, for the $k$-th class.

\begin{figure*}[!t]
  \centering
  \includegraphics[width=1.0\linewidth]{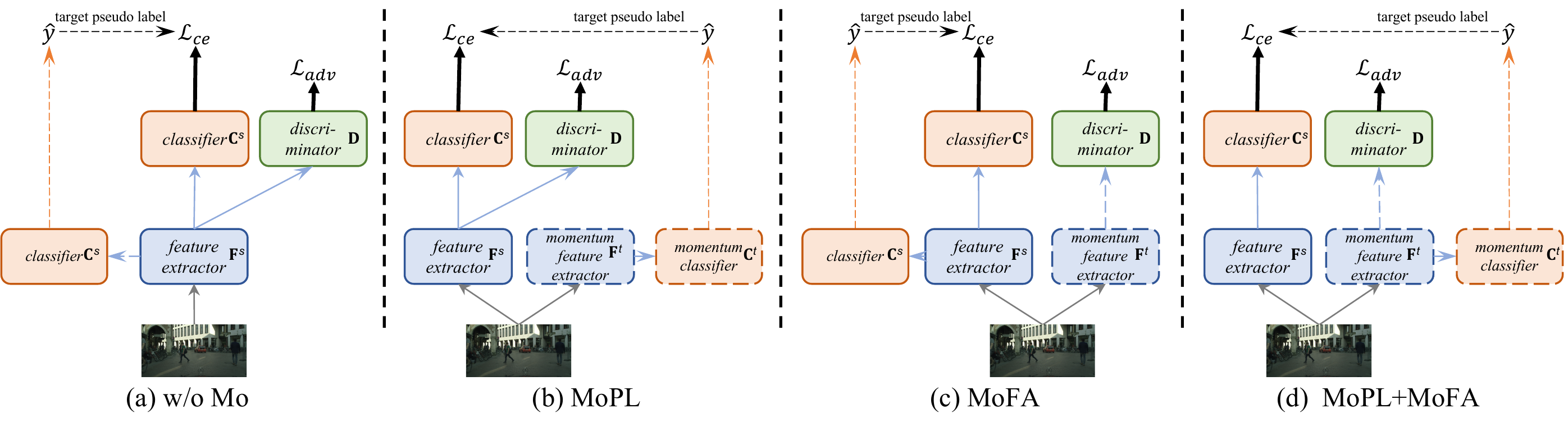}
  \vspace{-2em}
  \caption{ Smooth different supervisions for target domain in self-training and adversarial training. 
   Mo: Momentum Network, PL: for pseudo labels in self-training, FA: for features in adversarial training } 
  \label{fig:Mo}
\end{figure*}

Similarly, on target domain, we can get the construct the \emph{pseudo labels} $\hat{\vy}^t$ based on $f_{\theta(\cdot)}^t$.
We first obtain the categorical prediction on target domain:
\begin{equation}
\hat{\vp}^t = [\hat{p}^t_1,\dots,\hat{p}^t_{K^c}] = \mathbf{C}^t(\mathbf{F}^t(\vx^t)) = f_{\theta(\vx^t)}^t(\vx^t).
\label{eq.hatpt}
\end{equation}
Then, the target pseudo labels is defined as a one-hot vector:
\begin{align}
    \hat{y}_k &=
    \begin{cases}
    1, & \text{if }\ k=\arg \max_{k'} \hat{p}^t_{k'} 
    \\
    0, & \text{otherwise}.
    \end{cases}
\end{align}

To align the source and target domain and improve the discriminabilty of the latent space, we follow the common practice, \ie, \textbf{self-training} $f_{\theta(\cdot)}^s$ with both labeled source samples and pseudo-labeled target samples.
Following the above notations, that said we need to train $\mathbf{F}^s$ and $\mathbf{C}^s$ by $\mathcal{L}_{ce}^{target}$.
In this way, the networks $\mathbf{F}^s$ and $\mathbf{C}^s$ are bootstrapped by learning from pseudo labels that only get update till convergence, and then the updated labels are employed for the next training round. 
We obtain the segmentation categorical prediction on target domain by $f_{\theta(\vx^t)}^s$ as follows:
\begin{equation}
\aligned
    \vp^t &= [p^t_1,\dots,p^t_{K^c}] = \mathbf{C}^s(\mathbf{F}^s(\vx^t)) = f_{\theta(\vx^t)}^s(\vx^t)\\
\endaligned
\label{eq.pt}
\end{equation}
With the pseudo labels, the loss functions of segmentation in the target domain can then be defined as:
\begin{equation}
\aligned
    \mathcal{L}_{ce}^{target} &=  -\mathbb{E}_{\gD^t}[ \sum_{k=1}^{K^c} \hat{y}_{k}^t \log \left(p^t_{k}\right)], 
\endaligned
\label{eq.cet}
\end{equation}
where $p^t_k$ and $\hat{y}^t_k$ denote the predicted outputs and ground-truth, respectively, for the $k$-th class.

Traditional implementation by using adversarial training has been proven effective~\cite{hoffman2016fcns}. Particularly, it aligns features by using a binary domain discriminator $\mathbf{D}$~\cite{goodfellow2014generative}  to model the marginal distribution $\mathbb{P}(d|\mathbf{F}(\vx))$ by a $1$-dimensional vector. 
In contrast to perform the class-agnostic adversarial training,  \cite{wang2020classes,chen2017no} uses  a class-level domain discriminator to explicitly models the conditional distribution $\mathbb{P}(d,k|\mathbf{F}(\vx))$, according to $\mathbb{P}(d|\mathbf{F}(\vx))=\sum_{k=1}^{K^c}\mathbb{P}(d,k|\mathbf{F}(\vx))$, to  minimize the discrepancy between domains for each class separately. 
We will introduce the details of adversarial loss $\mathcal{L}_{adv}$ in \textsection~\ref{sec:Adaptation}.

\subsection{Momentum Network for Target Domain}
\label{sec:Momentum}

On target domain, we cannot directly train the segmentation network $f_{\theta(\cdot)}^t$ (constituted by feature extractor $\mathbf{F}^t$ and classifier $\mathbf{C}^t$), since there is no ground-truth.
A na\"ive solution is to use pseudo-labels and features generated by the source domain segmentation network $f_{\theta(\cdot)}^s$ to train $f_{\theta(\cdot)}^t$ in self-training and adversarial training.
As shown in Figure~\ref{fig:title}(a), target pseudo labels $\mathbf{C}^s(\mathbf{F}^s(\vx^t))$ are crucial for the classifier to make dense prediction, and target features  $\mathbf{F}^s(\vx^t)$ are essential for the discriminator to align features distribution.

\begin{figure*}[!t]
	\centering
	\includegraphics[width=1.0\linewidth]{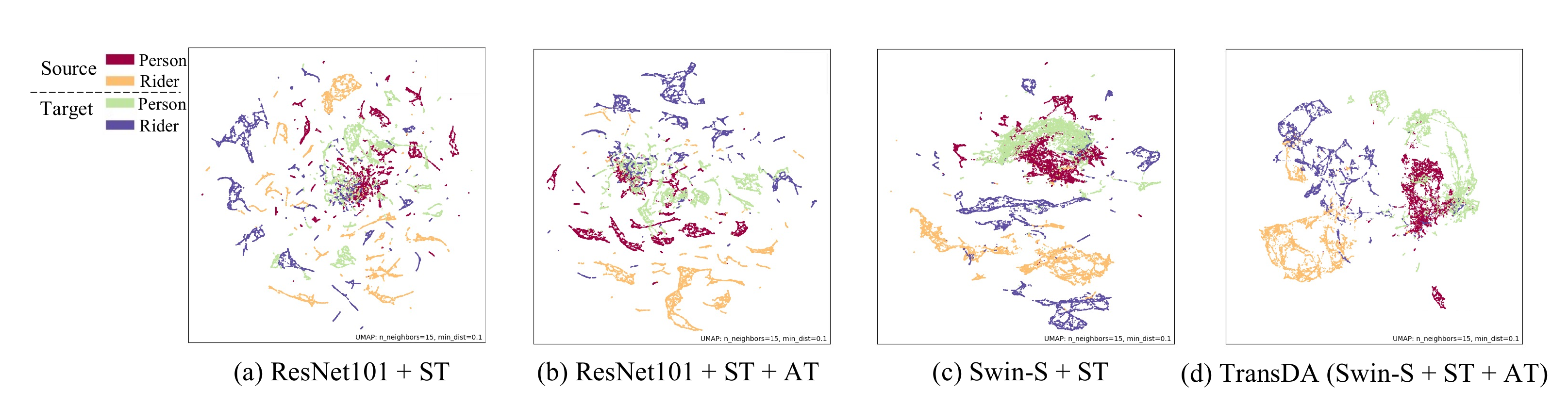} 
	\vspace{-2em}
	\caption{  The visualization of feature space, where we map features to 2D space with UMAP~\cite{mcinnes2018umap}. 
	  For a clear illustration, we only show two categories, \emph{i.e.}, red for source person, orange for source rider, green for target person, and blue for target rider. 
	Abbreviation: (S)elf (T)raining, (A)dversarial (T)raining.
	Data Augmentation is the default option and all ST+AT variants employ dynamic discrepancy and momentum network
	}
	\label{fig:umap}
\end{figure*}

However, through experiments, we find that such training schema on target domain is very unstable.
To further inspect target pseudo labels $\mathbf{C}^s(\mathbf{F}^s(\vx^t))$ and target features  $\mathbf{F}^s(\vx^t)$, we sample 8 images from Cityscapes train set~\cite{cordts2016cityscapes}, and track how drastically $\mathbf{D}(\mathbf{F}^s(\vx^t))$ and $\mathbf{C}^s(\mathbf{F}^s(\vx^t))$ change over time.
To be more specific, let's denote the output of these two functions at iteration $\ti$ as $\mathbf{D}_{\ti}(\mathbf{F}^s_\ti(\vx^t))$ and $\mathbf{C}^s_\ti(\mathbf{F}^s_\ti(\vx^t))$.
Then, we track both $\norm{\mathbf{D}_{\ti+1}(\mathbf{F}^s_{\ti+1}(\vx^t)) - \mathbf{D}_{\ti}(\mathbf{F}^s_\ti(\vx^t))}_1$ and 
$\norm{\mathbf{C}^s_{\ti+1}(\mathbf{F}^s_{\ti+1}(\vx^t)) - \mathbf{C}^s_\ti(\mathbf{F}^s_\ti(\vx^t))}_1$.
To make sure this phenomenon is consistent, we use $5$ random seeds to run the experiment, and the average and standard deviation of these distance sums are given in Figure~\ref{fig:title} (b) and (c) respectively.
As shown in Figure~\ref{fig:title}, by comparing the change of predictions under different backbones, we can see that the outputs of local ViTs (Swin-S) vibrate more drastically than conventional CNNs (ResNet-101), as indicated by the larger changes of  L1-distance over iterations.
We refer this problem as \emph{the high-frequency component problem in the predictions} of local ViTs (see Supplementary Material for details), and we argue that this problem causes that $f_{\theta(\cdot)}^s$ provides poor quality of supervision for $f_{\theta(\cdot)}^t$.

Formally, we can also formulate the learning of $f_{\theta(\cdot)}^t$ following the knowledge distillation \cite{hinton2015distilling} framework, where $f_{\theta(\cdot)}^s$ is the teacher model and $f_{\theta(\cdot)}^t$ is the student model.
Then, our above phenomenon is essentially the same to the ``zig-zag'' learning dynamics found in \cite{ren2021better}.
To be more specific, the authors find that change of teacher's predictions is very unsmooth, thus generates massive high-frequency components during its supervised learning.
Thus, they introduced a low-pass filter to smooth the predictions from teacher, and it turned out this can indeed improve the quality of supervision.
Although the teacher model and student model in the domain adaptation task are from \emph{two different domains}, in contrast to single domain in \cite{ren2021better}, we argue that the above issues can also be tackled by  a low-pass filter.
However, the original implementation in  \cite{ren2021better}, \ie, a look-up table which stores predictions for all training example, is very infeasible in semantic segmentation as it requires huge storage overhead for all pixels. With further analysis, the drastic change of predictions actually indicate that the parameters are changing drastically.
Thus, we argue that momentum network can also be used to smooth the parameters of  $\mathbf{F}^t$ and $\mathbf{C}^t$ in order to smooth the supervision provided by $f_{\theta(\cdot)}^s$  illustrated in the above paragraph.

As illustrated in Figure~\ref{fig:Mo}, MoPL only smooths the target pseudo label supervision $\mathbf{C}^s(\mathbf{F}^s(\vx^t))$ for source classifier $\mathbf{C}^s$ and feature extractor $\mathbf{F}^s$, MoFA only smooths the target feature supervision $\mathbf{F}^s(\vx^t)$ for discriminator $\mathbf{D}$, and MoPL+MoFA jointly smooth the $\mathbf{C}^s(\mathbf{F}^s(\vx^t))$ and $\mathbf{F}^s(\vx^t)$.
As shown in Table~\ref{tab:Ablation}, we find that MoPL+MoFA is the only way for local ViT model to succeed in this task, revealing that jointly smoothing the pseudo labels and features is essential.
Concretely, the feature extractor $\mathbf{F}^t$ and classifier $\mathbf{C}^t$ on target domain are updated by a momentum copy from source domain $f_{\theta(\cdot)}^s$ in Eqn.(\ref{eq.new-problem}). 
Note that only the parameters $\vartheta^s$ (of $\mathbf{F}^s$ and $\mathbf{C}^s$) are updated by the gradient of $\mathcal{L}_{ce}^{source}$,  $\mathcal{L}_{ce}^{target}$ and $\mathcal{L}_{adv}$.
Meanwhile, since $f_{\theta(\cdot)}^t$ is the momentum network of $f_{\theta(\cdot)}^s$, the gradient cannot directly update $f_{\theta(\cdot)}^t$. 
As shown in \textsection~\ref{sec:ablation momentum}, we find that a relatively large momentum (\eg, \mbox{$m$ $=$ 0.999}, our default) works much better than a smaller value (\eg, \mbox{$m$ $=$ $0.9$}), suggesting that a slowly evolving $f_{\theta(\cdot)}^t$ can effectively smooth the target pseudo labels and features (Figure~\ref{fig:title}(b) and (c)).

\newcommand{\tabincell}[2]{\begin{tabular}{@{}#1@{}}#2\end{tabular}}  
\begin{table*}[!t]
\centering
\footnotesize
\resizebox{0.7\textwidth}{!}{
\begin{tabular}{c|ccc|cc | cc} 
\toprule 
\multirow{2}{*}{adversarial} &\multirow{2}{*}{\tabincell{c}{dynamic\\ discrepancy}}&\multirow{2}{*}{MoPL}&\multirow{2}{*}{MoFA}&\multicolumn{2}{c|}{\textbf{ResNet-101}(43M)} & \multicolumn{2}{c}{\textbf{Swin-S}(50M)} 
\\
   & & &  &mIoU & gain & mIoU & gain\\
\midrule
 \multirow{2}{*}{\tabincell{c}{w/o\\discriminator}}&&&&\cellcolor{pink}$46.7_{\pm0.8}$&\cellcolor{pink}-&\cellcolor{pink}$52.5_{\pm5.1}$&\cellcolor{pink}-\\
&&\checkmark&&\cellcolor{pink}$46.3_{\pm0.7}$&\cellcolor{pink}$\downarrow0.4$&\cellcolor{pink}$47.4_{\pm0.2}$&\cellcolor{pink}$\downarrow5.1$\\
\midrule
\multirow{8}{*}{\tabincell{c}{binary\\discriminator}}&&&&\cellcolor{lime}$48.8_{\pm0.5}$&\cellcolor{lime}$\uparrow{2.1}$&\cellcolor{lime}${53.9}_{\pm2.3}$&\cellcolor{lime}$\uparrow{1.4}$\\
&&\underline{\checkmark}&&\cellcolor{lime}\underline{${49.0}_{\pm0.3}$}&\cellcolor{lime}\underline{$\uparrow2.3$}&\cellcolor{pink}$51.8_{\pm1.3}$&\cellcolor{pink}$\downarrow0.7$\\
&&&\checkmark&\cellcolor{pink}$45.9_{\pm0.5}$&\cellcolor{pink}$\downarrow0.8$&\cellcolor{pink}$46.8_{\pm1.7}$&\cellcolor{pink}$\downarrow5.7$\\
&&\checkmark&\checkmark&\cellcolor{pink}$46.4_{\pm0.4}$&\cellcolor{pink}$\downarrow0.3$&\cellcolor{lime}${58.9}_{\pm0.5}$&\cellcolor{lime}$\uparrow{6.4}$\\
&\checkmark&&&\cellcolor{lime}${47.9}_{\pm0.3}$&\cellcolor{lime}$\uparrow{1.2}$&\cellcolor{lime}${54.5}_{\pm1.3}$&\cellcolor{lime}$\uparrow{1.2}$\\
&\checkmark&\checkmark&&\cellcolor{lime}${48.6}_{\pm0.2}$&\cellcolor{lime}$\uparrow{1.9}$&\cellcolor{lime}$52.6_{\pm0.1}$&\cellcolor{lime}$\uparrow0.1$\\
&\checkmark&&\checkmark&\cellcolor{pink}$46.3_{\pm0.9}$&\cellcolor{pink}$\downarrow0.4$&\cellcolor{pink}$49.2_{\pm3.2}$&\cellcolor{pink}$\downarrow3.3$\\
&\underline{\checkmark}&\underline{\checkmark}&\underline{\checkmark}&\cellcolor{lime}$46.7_{\pm0.7}$&\cellcolor{lime}$\uparrow0.0$&\cellcolor{lime}\underline{${59.3}_{\pm0.7}$}&\cellcolor{lime}\underline{$\uparrow{6.8}$}\\

\midrule
 \multirow{8}{*}{\tabincell{c}{class\\discriminator}}&&&&\cellcolor{lime}${47.9}_{\pm1.1}$&\cellcolor{lime}$\uparrow{1.2}$&\cellcolor{lime}$52.7_{\pm1.1}$&\cellcolor{lime}$\uparrow0.2$\\
&&\underline{\checkmark}&&\cellcolor{lime}$\underline{{49.6}}_{\pm0.5}$&\cellcolor{lime}\underline{$\uparrow{2.9}$}&\cellcolor{pink}$49.3_{\pm1.1}$&\cellcolor{pink}$\downarrow3.2$\\
&&&\checkmark&\cellcolor{pink}$45.3_{\pm1.5}$&\cellcolor{pink}$\downarrow1.4$&\cellcolor{pink}$49.0_{\pm4.6}$&\cellcolor{pink}$\downarrow3.5$\\
&&\checkmark&\checkmark&\cellcolor{pink}$46.4_{\pm0.5}$&\cellcolor{pink}$\downarrow0.3$&\cellcolor{lime}${57.2}_{\pm0.8}$&\cellcolor{lime}$\uparrow{4.7}$\\
&\checkmark&&&\cellcolor{lime}${47.4}_{\pm0.9}$&\cellcolor{lime}$\uparrow{0.7}$&\cellcolor{lime}${55.5}_{\pm1.6}$&\cellcolor{lime}$\uparrow{3.0}$\\
&\checkmark&\checkmark&&\cellcolor{lime}${48.5}_{\pm0.5}$&\cellcolor{lime}$\uparrow{1.8}$&\cellcolor{pink}$49.4_{\pm1.4}$&\cellcolor{pink}$\downarrow3.1$\\
&\checkmark&&\checkmark&\cellcolor{pink}$45.2_{\pm1.3}$&\cellcolor{pink}$\downarrow1.5$&\cellcolor{pink}$48.4_{\pm3.4}$&\cellcolor{pink}$\downarrow4.1$\\
&\underline{\checkmark}&\underline{\checkmark}&\underline{\checkmark}&\cellcolor{lime}$46.8_{\pm0.3}$&\cellcolor{lime}$\uparrow0.1$&\cellcolor{lime}\underline{${58.2}_{\pm0.9}$}&\cellcolor{lime}\underline{$\uparrow{5.7}$}\\
\bottomrule 
\end{tabular}
 }
\vspace{-1em}
\caption { Smoothing matters on different backbones. 
 All variants employ self-training on GTA5$\to$Cityscapes. 
 MoPL:  Momentum Network for target pseudo labels, MoFA: Momentum Network for target features in adversarial training
 }
\label{tab:Ablation}%
\end{table*}%

\subsection{Dynamic of Discrepancy Measurement}
\label{sec:Adaptation}

This subsection presents how to realize the  dynamic  of  discrepancy  measurement $\text{Dis}_{w(\vx^s,\vx^t)}$ in Eq.~\ref{eq.new-problem}. 
To avoid the exacerbation by the noise and errors, we propose to use \emph{dynamic weighting} $w$ to suppress the influence of false samples in the transfer process. Concretely, the larger $w$ is, the more reliable transferability it is. 
For binary domain discriminator, we realize the dynamic discrepancy $\text{Dis}_{w(\vx^s,\vx^t)}$ by the weighted adversarial loss below:
\begin{equation}
\aligned
\mathcal{L}_{adv} &=  -\mathbb{E}_{\gD^s} [w^s(\vx^s)\log \mathbf{D}(\mathbf{F}^s(\vx^s)))]\\
                & -\mathbb{E}_{\gD^t} [w^t(\vx^t)\log (1-\mathbf{D}(\mathbf{F}^t(\vx^t)))],
\endaligned
\label{eq.cadv}
\end{equation}
where the dynamic weighting $w(\vx^s, \vx^t)$ is implemented as $w^s(\vx^s)$ for the source input and $w^t(\vx^t)$ for the target input.
For class-level domain discriminator, please refer to Supplementary Material.

Similiar to UniDA~\cite{you2019universal}, we use a non-adversarial domain similarity network $\mathbf{S}$ to obtain the domain similarity to the source domain. The objective of $\mathbf{S}$ is to predict the samples from source domain as 1 and those from target domain as 0. The loss function of domain similarity is given as follows:
\begin{equation}
\aligned
    \mathcal{L}_{sim} &= -\mathbb{E}_{\gD^s} \log \mathbf{S}(\mathbf{F}^s(\vx^s)) -\mathbb{E}_{\gD^t} \log (1-\mathbf{S}(\mathbf{F}^t(\vx^t))).
\endaligned
\label{eq.sim}
\end{equation}
Here, $\mathbf{S}(\cdot) \in [0,1]$ can be seen as the quantification for the domain similarity of each sample. 
For a source sample, smaller score means that it is more similar to the target domain; for a target sample,  vice versa.
Therefore, we can hypothesize that $\mathbb{E}_{x\sim \gD_{\gY^{sp}}}\mathbf{S}(\mathbf{F}^s(\vx^{\gY^{sp}})) >  \mathbb{E}_{x\sim \gD_{\gY^c}}\mathbf{S}(\mathbf{F}(\vx^{\gY^c})) > \mathbb{E}_{x\sim \gD_{\gY^{tp}}}\mathbf{S}(\mathbf{F}^t(\vx^{\gY^{tp}}))$.

We now depict how to obtain the weights $w(\vx^s)$  and $w(\vx^t)$ in Eq.~\ref{eq.cadv}. Given each input $\vx$, its domain similarity $\mathbf{S}(\vx)$ and segmentation categorical prediction $\vp(\vx)$ over the label set $\gY^c$, we dynamically compute $w^s(x)$ and $w^t(x)$  by:

\begin{equation}
\label{eq.weight}
\begin{split}
    w^s(\vx^s) &= \frac{E(\vp^s)}{\log |K^c| } - \mathbf{S}(\mathbf{F}^s(\vx^s)),\\
    w^t(\vx^t) &= \mathbf{S}(\mathbf{F}^t(\vx^t)) - \frac{E(\hat{\vp}^t)}{\log |K^c| }.
\end{split}
\end{equation}
Note that the entropy $E(\vp)$ is normalized by its maximum value ($\log |K^c|$) so that it is restricted into [0, 1] and comparable to the domain similarity $\mathbf{S}(\mathbf{F}(\vx))$.  The entropy $E(\vp(\vx))$ can quantify the uncertainty, and smaller entropy means more confident prediction. Thus, we can assume
that $\mathbb{E}_{x\sim \gD_{\gY^{tp}}}E(\vp^{\gY^{tp}}) > \mathbb{E}_{x\sim \gD_{\gY^c}}E(\vp^{\gY^c}) > \mathbb{E}_{x\sim \gD_{\gY^{sp}}}E(\vp^{\gY^{sp}})$.
Besides, the weights $w$ are min-max normalized into interval [0, 1].
The motivation is that for any source sample, it should be weighted more, if there is larger uncertainty within the prediction and smaller similarity to the source domain. For the target sample, vice versa.

\subsection{Full Training Procedure}
\label{sec:Training}

By assembling the above modules, we propose a simple training schedule for TransDA to joint self-training along with adversarial training.
The training process is proceeded in terms of curriculum-like learning.
\paragraph{Warm-up Phase:} We load and fine-tune the Swin model in terms of  cross-entropy (CE) loss (Eq.~\ref{eq.ces}) in source domain, the domain similarity loss (Eq.~\ref{eq.sim})and the weighted adversarial loss (Eq.~\ref{eq.cadv}). The full objective of warm-up phase:
\begin{equation}
\aligned
	\min_{\mathbf{F},\mathbf{C},\mathbf{S}} \max_{\mathbf{D}} \mathcal{L}_{ce}^{source}   + \mathcal{L}_{sim} - \mathcal{L}^{}_{adv}.
\endaligned
\end{equation}

\paragraph{Train Phase:}
We further transfer knowledge from the warm-up phase segmentation network to a pretrained student network (\emph{i.e.}, Swin pretrained on ImageNet), by the self-training paradigm~\cite{zhang2021prototypical}. The teacher model generates one-hot pseudo labels $\bm{\hat{y}}$ to teach the student network via a cross-entropy loss.
It can be carried out for several rounds, and for each round, the student model is always initialized as the pretrained model in order to escape from the local optima in the last round. We conduct self-training in terms of cross-entropy (CE) loss (Eq.~\ref{eq.ces} and~\ref{eq.cet}), the domain similarity loss (Eq.~\ref{eq.sim})and the weighted adversarial loss (Eq.~\ref{eq.cadv}). The full objective of start phase is as follows:
\begin{equation}
\aligned
	\min_{\mathbf{F},\mathbf{C},\mathbf{S}} \max_{\mathbf{D}} \mathcal{L}_{ce}^{source} + \mathcal{L}_{ce}^{target}   + \mathcal{L}_{sim} - \mathcal{L}^{}_{adv}.
\endaligned
\end{equation}

\begin{figure*}[!t]
	\centering
	\includegraphics[width=0.9\linewidth]{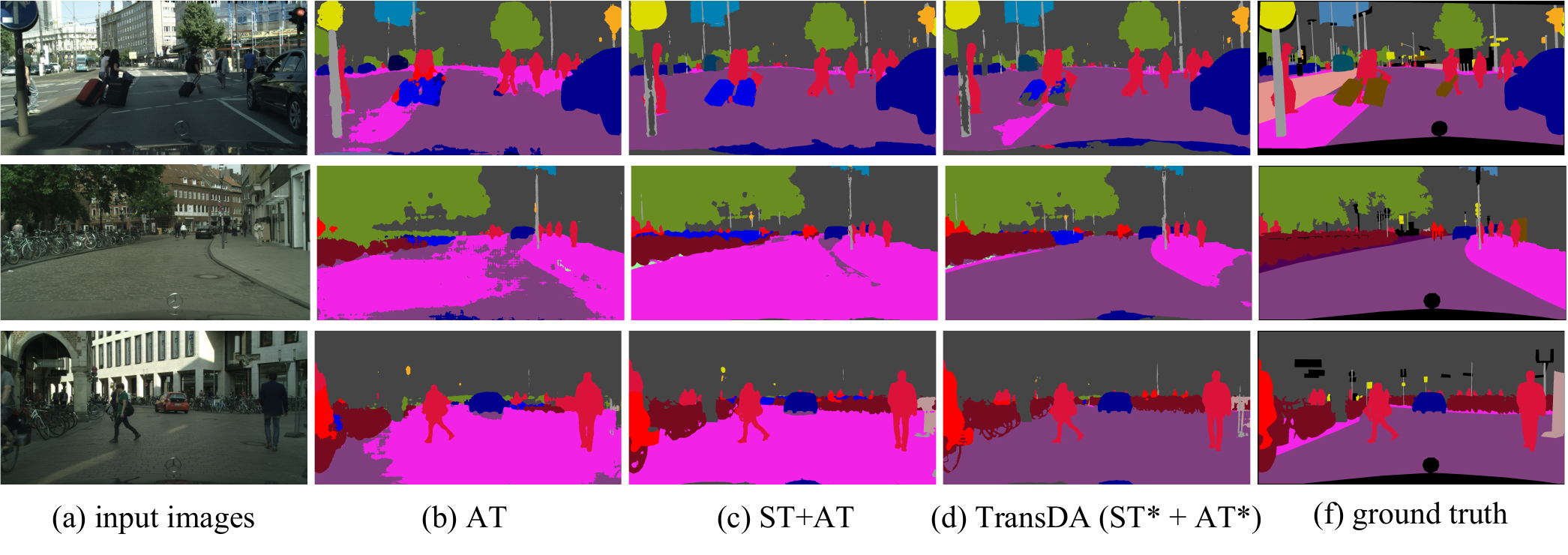} 
	\vspace{-1.5em}
	\caption{Visualizations of domain adaptive semantic segmentation based on Swin ViT at the inference stage.   Abbreviation: (S)elf (T)raining, (A)dversarial (T)raining.
	* denotes employing our proposed smoothing method.
	}
	\label{fig:qualitative}
\end{figure*}

\begin{table*}[!t]
\centering
\footnotesize
\resizebox{1.0\textwidth}{!}{
\begin{tabular}{@{}c|llllll@{}}
\toprule 
 comp.  & vanilla   &  DA & DA+AT & DA+ST & DA+ST+AT & DA+ST$^{*}$+AT$^{*}$
\\
\midrule
mIoU & $38.4_{\pm 2.7}$ & $45.9_{\pm 2.7}(\color{green}{\uparrow}\color{black}{7.5})$ & $49.7_{\pm 1.2}(\color{green}{\uparrow}\color{black}{11.3})$ & $52.5_{\pm 5.1}(\color{green}{\uparrow}\color{black}{14.1})$ 
&  $53.9_{\pm 2.3}(\color{green}{\uparrow}\color{black}{15.5}) $
& $59.3_{\pm 0.7}(\color{green}{\uparrow}\color{black}{20.9}) $
\\
\bottomrule  
\end{tabular}%
}
\vspace{-1em}
\caption {Foundation components in this new local ViT-backbone baseline. 
}
\label{tab:Component}%
\end{table*}%

\section{Experiments}
\subsection{Implementation}
\paragraph{Datasets.}
We evaluate the performance of our methods on the challenging  domain adaptive semantic segmentation task.
The source domain contains two synthetic datasets, GTA5 \cite{richter2016playing} and SYNTHIA\cite{ros2016synthia} and the target domain is a real dataset: Cityscapes \cite{cordts2016cityscapes}. We conduct experiments on two domain adaption flows: GTA5$\rightarrow$Cityscapes and SYNTHIA$\rightarrow$Cityscapes.
To be specific,  GTA5 and SYNTHIA share 19 and 16 common categories with Cityscapes, repectively. 
On SYNTHIA$\rightarrow$Cityscapes, following\cite{ma2021coarse}, we consider two different testing protocols: applying all 16 common categories or just a subset consisting of 13 categories  for evaluations. Note that we train the model on the whole training set for both settings. Readers can refer to Supplementary Material for more datasets details.

\paragraph{Setup.}

We leverage Swin Transformer~\cite{liu2021Swin} as the feature extractor $\mathbf{F}$. Specifically, to ensure the fair comparison to the classical backbone ResNet-101, we load the Swin-S model which is pre-trained on ImageNet-1K and has the similar model size and computation complexity to ResNet-101 as the backbone and denote our model as TransDA-S. Meanwhile, we also employ a larger model Swin-B which is pre-trained on ImageNet-22K with the input size of 224 $\times$ 224 and denote it as TransDA-B. 
Moreover, we utilize the ubiquitous data augmentation techniques~\cite{wang2020classes,araslanov2021self} to enhance the training stability, including random horizontal flipping, random re-scaling within ratio range $[0.5, 2.0]$,  color jittering with brightness, contrast, saturation, and hue. We use the single-scale test at the inference stage. 
Please refer to Supplementary Material for more training details.

\paragraph{Metrics.}
Following the common protocol in this area, we use PASCAL VOC Intersection-over-Union (IoU) \cite{everingham2015pascal} as the evaluation metric. 
Notably, previous methods benchmark the performance by only reporting the best numbers without reflecting the averaged scores and standard deviations. 
In our experiments, we recommend to show the robustness by further reporting the averaged scores and standard deviations. We report the best value in Table~\ref{tab:gta5_to_city}  and Table~\ref{tab:synthia_to_city} and report the averaged scores and standard deviations in Supplementary Material.

\begin{table*}[!t]
\centering
\resizebox{\textwidth}{!}{
\begin{tabular}{c|*{19}{c}|c}
\toprule 
  & \multicolumn{1}{c}{\begin{turn}{60}road\end{turn}} & \multicolumn{1}{c}{\begin{turn}{60}sideway\end{turn}} & \multicolumn{1}{c}{\begin{turn}{60}building\end{turn}} & \multicolumn{1}{c}{\begin{turn}{60}wall\end{turn}} & \multicolumn{1}{c}{\begin{turn}{60}fence\end{turn}} & \multicolumn{1}{c}{\begin{turn}{60}pole\end{turn}} & \multicolumn{1}{c}{\begin{turn}{60}light\end{turn}} & \multicolumn{1}{c}{\begin{turn}{60}sign\end{turn}} & \multicolumn{1}{c}{\begin{turn}{60}vege.\end{turn}} & \multicolumn{1}{c}{\begin{turn}{60}terrace\end{turn}} & \multicolumn{1}{c}{\begin{turn}{60}sky\end{turn}} & \multicolumn{1}{c}{\begin{turn}{60}person\end{turn}} & \multicolumn{1}{c}{\begin{turn}{60}rider\end{turn}} & \multicolumn{1}{c}{\begin{turn}{60}car\end{turn}} & \multicolumn{1}{c}{\begin{turn}{60}truck\end{turn}} & \multicolumn{1}{c}{\begin{turn}{60}bus\end{turn}} & \multicolumn{1}{c}{\begin{turn}{60}train\end{turn}} & \multicolumn{1}{c}{\begin{turn}{60}motor\end{turn}} & \multicolumn{1}{c}{\begin{turn}{60}bike\end{turn}} & \multicolumn{1}{|c}{mIoU}\\
\midrule
\multicolumn{20}{l}{\emph{Backbone: ResNet-101 (43M)}}\\
\midrule 
{w/o  Adaptation~\cite{wang2020classes}} &65.0&16.1&68.7&18.6&16.8&21.3&31.4&11.2&83.0&22.0&78.0&54.4&33.8&73.9&12.7&30.7&{13.7}&28.1&19.7&36.8\\
{AdaptSeg~\cite{tsai2018learning} }&86.5 & 25.9 & 79.8 & 22.1& 20.0& 23.6& 33.1& 21.8& 81.8& 25.9& 75.9& 57.3& 26.2& 76.3& 29.8& 32.1& 7.2& 29.5& 32.5& 41.4\\
{ Seg-Uncertainty~\cite{zheng2021rectifying}} &90.4& 31.2& 85.1& 36.9& 25.6& 37.5& 48.8& 48.5& 85.3& 34.8& 81.1& 64.4& 36.8& 86.3& 34.9& 52.2& 1.7& 29.0& 44.6& 50.3\\
{ MetaCorrection~\cite{guo2021metacorrection}}&{92.8}&58.1&86.2&39.7&33.1&36.3&42.0&38.6&85.5&37.8&87.6&62.8&31.7&84.8&35.7&50.3&2.0&36.8&48.0&52.1\\
{SAC~\cite{araslanov2021self}}& 90.4 & 53.9 &  86.6 &  42.4 & 27.3 &  45.1 & 48.5 &  42.7 &  87.4 & 40.1 & 86.1 &  67.5 & 29.7 &  88.5 &  49.1 &  54.6 & 9.8 & 26.6 & 45.3 &  53.8 \\
{ Coarse-to-Fine~\cite{ma2021coarse}} & 92.5 & {58.3} & 86.5 & 27.4 & 28.8 & 38.1 & 46.7 & 42.5 & 85.4 & 38.4 & 91.8 & 66.4 & 37.0 & 87.8 & 40.7 & 52.4 & \textbf{44.6} & 41.7 & \textbf{59.0} & 56.1\\
{ProDA~\cite{zhang2021prototypical}}  & 87.8 & 56.0  & 79.7  & \textbf{46.3} & \textbf{44.8}  & 45.6  &53.5  & \textbf{53.5} & 88.6  & \textbf{45.2}  & 82.1 & 70.7 & 39.2  & 88.8  & 45.5  & \textbf{59.4}  & 1.0  & \textbf{48.9} & 56.4  &  57.5\\
\midrule 
\multicolumn{21}{l}{\emph{Backbone: Swin-S ViT (50M)}}\\
\midrule 
{w/o Adaptation}  &55.9&21.8&63.1&14.0&22.0&27.2&46.8&17.4&83.3&32.8&86.1&62.2&28.7&43.8&32.2&36.9&1.1&34.8&35.5&39.2\\
{TransDA-S}  & \textbf{92.9}& \textbf{59.1}& \textbf{88.2}& {42.5}& {32.0}& \textbf{47.6}& \textbf{57.6}& {39.2}& \textbf{89.6}& {42.0}& \textbf{94.1}& \textbf{74.3}& \textbf{45.3}& \textbf{91.4}& \textbf{54.0}& {58.0}& {44.4}& {48.3}& {51.4} &  \textbf{60.6}\\
\midrule 
\midrule 
\multicolumn{21}{l}{\emph{Backbone: Swin-B ViT (88M)}}\\
\midrule 
{w/o Adaptation} &63.3&28.6&68.3&16.8&23.4&37.8&51.0&34.3&83.8&42.1&85.7&68.5&25.4&83.5&36.3&17.7&2.9&36.1&42.3&44.6\\
{TransDA-B}  & {94.7}& {64.2}& {89.2}& 48.1& 45.8& 50.1& 60.2& 40.8& {90.4}& {50.2}& {93.7}& 76.7& 47.6& 92.5& 56.8& 60.1& {47.6}& {49.6}& {55.4}& {63.9}\\
\bottomrule  
\end{tabular}%
}
\vspace{-.5em}
\caption { Comparisons with state-of-the-art methods on GTA5$\to$Cityscapes}
\label{tab:gta5_to_city}%
\end{table*}%

\begin{table*}[!h]
\centering
\resizebox{\textwidth}{!}{
\begin{tabular}{@{}c|*{16}{c}|c|c@{}}
\toprule
 & \multicolumn{1}{c}{\begin{turn}{60}road\end{turn}} & \multicolumn{1}{c}{\begin{turn}{60}sideway\end{turn}} & \multicolumn{1}{c}{\begin{turn}{60}building\end{turn}} & \multicolumn{1}{c}{\begin{turn}{60}wall*\end{turn}} & \multicolumn{1}{c}{\begin{turn}{60}fence*\end{turn}} & \multicolumn{1}{c}{\begin{turn}{60}pole*\end{turn}} & \multicolumn{1}{c}{\begin{turn}{60}light\end{turn}} & \multicolumn{1}{c}{\begin{turn}{60}sign\end{turn}} & \multicolumn{1}{c}{\begin{turn}{60}vege.\end{turn}} & \multicolumn{1}{c}{\begin{turn}{60}sky\end{turn}} & \multicolumn{1}{c}{\begin{turn}{60}person\end{turn}} & \multicolumn{1}{c}{\begin{turn}{60}rider\end{turn}} & \multicolumn{1}{c}{\begin{turn}{60}car\end{turn}}& \multicolumn{1}{c}{\begin{turn}{60}bus\end{turn}} & \multicolumn{1}{c}{\begin{turn}{60}motor\end{turn}} & \multicolumn{1}{c}{\begin{turn}{60}bike\end{turn}}& \multicolumn{1}{|l}{mIoU} & \multicolumn{1}{|l}{mIoU*}\\
\midrule
\multicolumn{19}{l}{\emph{Backbone: ResNet-101 (43M)}}\\
\midrule 
 w/o Adaptation~\cite{wang2020classes} & 55.6& 23.8& 74.6& 9.2& 0.2& 24.4& 6.1& 12.1& 74.8& 79.0& 55.3& 19.1& 39.6& 23.3& 13.7& 25.0& 33.5& 38.6\\
{AdaptSeg~\cite{tsai2018learning}} & 79.2& 37.2& 78.8& -& -& -& 9.9& 10.5& 78.2& 80.5& 53.5& 19.6& 67.0& 29.5& 21.6& 31.3&-  &45.9 \\
{Seg-Uncertainty~\cite{zheng2021rectifying}} & 87.6& 41.9& 83.1& 14.7& {1.7}& 36.2& 31.3& 19.9& 81.6& 80.6& 63.0& 21.8& 86.2& 40.7& 23.6& {53.1} & 47.9 &  54.9 \\
{MetaCorrection~\cite{guo2021metacorrection}}&\textbf{92.6}&\textbf{52.7}&{81.3}&8.9&\textbf{2.4}&28.1&13.0&7.3&83.5&{85.0}&60.1&19.7&84.8&{37.2}&21.5&43.9&45.1&{52.5} \\
{SAC~\cite{araslanov2021self}}&  89.3 & 47.2 &  85.5 &  26.5 & 1.3 &  43.0 &  45.5 & 32.0 &  87.1 &  89.3 & 63.6 & 25.4 & 86.9 & 35.6 & 30.4 & 53.0 &  52.6&  59.3  \\
{Coarse-to-Fine~\cite{ma2021coarse}} & 75.7 & 30.0 & 81.9 & 11.5 & 2.5 & 35.3 & 18.0 & 32.7 & 86.2 & 90.1 & 65.1 & \textbf{33.2} & 83.3 & 36.5 & 35.3 & \textbf{54.3} & 48.2 & 55.5 \\
{ProDA~\cite{zhang2021prototypical}}  & {87.8} &   45.7 & {84.6} &  \textbf{37.1} & 0.6 &  {44.0} &  \textbf{54.6} &  \textbf{37.0} &  {88.1} &  84.4 & \textbf{74.2}  &  24.3 &  {88.2} &  {51.1} &  {40.5} &  45.6 & \textbf{55.5} & {62.0} \\
\midrule 
\multicolumn{19}{l}{\emph{Backbone: Swin-S ViT (50M)}}\\
\midrule 

{w/o Adaptation}  &30.6&26.1&42.9&3.8&0.1&25.9&32.3&15.6&80.3&70.7&60.5&8.2&69.0&30.3&11.2&12.3&32.5&37.7\\
{TransDA-S}  & 82.1& 40.9& \textbf{86.2}& 25.8& 1.0& \textbf{53.0}& 53.7& 36.1& \textbf{89.2}& \textbf{90.3}& 68.0& 26.2& \textbf{90.9}& \textbf{58.4}& \textbf{41.2}& 45.4 & \textbf{55.5} & \textbf{62.2} \\
\midrule 
\midrule 
\multicolumn{19}{l}{\emph{Backbone: Swin-B ViT (88M)}}\\
\midrule 
{w/o Adaptation} &57.3&33.8&56.0&6.3&0.2&33.8&35.5&18.9&79.9&74.8&63.1&10.9&78.3&39.0&20.8&19.4&39.2&45.2\\
{TransDA-B}  & 90.4 & {54.8} & {86.4} & 31.1 & 1.7 & {53.8} & {61.1} & {37.1} & {90.3} & 93.0 & 71.2 & 25.3 & {92.3} & {66.0} & 44.4& 49.8& {59.3}&   {66.3} \\
\bottomrule  
\end{tabular}%
}
\vspace{-.5em}
\caption {Comparisons with state-of-the-art methods on SYNTHIA$\to$Cityscapes.  
  mIoU and mIoU* denote the scores across 16 and 13 categories respectively}
\label{tab:synthia_to_city}%
\end{table*}%

\subsection{Ablation study: smoothing matters}
\label{sec:ablation smoothing}

To validate the effectiveness of our proposed key techniques:momentum network and dynamic of discrepancy measurement, we conduct the ablation study under different backbones on the GTA5$\rightarrow$Cityscapes. We report the results in Table~\ref{tab:Ablation} and have the following observations:
\begin{itemize}[leftmargin=*]
\item As can be seen in the third row of results, the gain of backbone replacement is +5.1 (53.9-48.8) when employing just the almost ubiquitous self-training, which validates the cross-domain transferability of local ViT backbone. However, by addressing the high-frequency components problem in local ViT, our proposed method can further boost the performance +5.4 (59.3-53.9). This additional improvement can be viewed as significant since it is much larger than that of AdaptSeg~\cite{tsai2018learning} (+3.1), which is the first work to replace VGG with ResNet in this task. 
\item We also observe different trends for different models under the same setting, such as row 2 and 4 in binary discriminator and row 2,4 and 6 in class discriminator. It implies the inherent difference between two backbones, which justifies the motivation of our model design.
\item On Swin-S, the proposed momentum network dramatically improves the performance (+5.0=58.9-53.9 on binary discriminator). It indicates that Swin-S suffers from the severe high-frequency component problem both in target pseudo labels and features, it is vital to smooth both at the same time. 
\item For ResNet-101, smoothing pseudo labels can also improve its performance, while smoothing both pseudo labels and target features or applying dynamic discrepancy will degenerate its performance. This may due to that the target features of ResNet-101 are already smooth enough (shown in Figure~\ref{fig:title}) and the different feature representations learnt by CNNs (illustrated in Figure~\ref{fig:umap}). Adding the additional smoothing would damage its target features.
\end{itemize}

\subsection{Ablation study: foundation components}
\label{sec:ablation foundation}
As the first work to build a new baseline based on a local ViT backbone, it is essential to ablate the foundation components to provide the basic insight.

As shown in Table~\ref{tab:Component} and Figure~\ref{fig:qualitative}, we explore the potential of leveraging local ViT backbones via a classical training paradigm. 
We observe: i) Besides in fully supervised learning, data augmentation techniques also play an vital role in local ViT backbones for domain adaptation; ii) Adversarial training can further improve the transferability, but the effect is not as obvious as using self-training alone; iii) If they are adopted jointly, a bottleneck for further improvement is encountered. However, with our proposed smoothing method, the potential of the model under this classical training paradigm is fully released (+5.4=59.3-53.9).

To better develop intuition, we visualize the learned features for TransDA in Figure~\ref{fig:umap}.   We find that: 
i) By using ResNet-101, the predictions of different classes are mixed together in both domains, showing the weak ability in semantic distinction;
ii) When applying self-training upon Swin-S, the samples of different classes are well separated, which is probably owing to the better expressivity of local ViT. Meanwhile, particularly for the class ``person", the source domain and target domain overlap with each other by a large percentage, showing the better generalization ability across domains;
iii) After applying adversarial training upon Swin-S, 
the shape of the feature space is closer to a sphere.
At the same time, the samples of both domains are well separated across different categories.

Compared with conventional CNNs, the representations of local ViTs are significantly different. 
Taking the interesting inspiration from~\cite{han2021demystifying} that the attentions used in local ViTs are indeed equivalent to depth-wise CNNs with dynamic weight, local ViTs acquire the \emph{dynamic} nature of $f_{\theta(\cdot)}$. Joining the study by DRT~\cite{li2021dynamic}, the dynamic nature may break down domain barriers better. 
Based on the above observations and discussions, it reveals that it's an urge to explore the original techniques in this field to develop new methods based on this new backbone to break through the performance ceiling.

\subsection{Ablation study: momentum coefficient}
\label{sec:ablation momentum}
The table below shows TransDA-S mIoU on GTA5$\to$Cityscapes with different momentum values ($m$ in Eqn.(\ref{eq.new-problem})):
\begin{center}
\resizebox{0.475\textwidth}{!}{	
\begin{tabular}{c|ccccc}
momentum $m$ & 0 & 0.9 & 0.99 & 0.999 & 0.9999 \\
\hline
mIoU & $54.5_{\pm 1.3}$ & $51.6_{\pm 2.3}$ & $56.5_{\pm 2.0}$ & $59.3_{\pm 0.7}$ & $53.1_{\pm 0.4}$ \\
\end{tabular}}
\end{center}
It performs reasonably well when $m$ is 0.99 and 0.999, showing that a smooth (\ie, relatively large momentum) $\mathbf{F}^t$ and $\mathbf{C}^t$ is beneficial.
When $m$ is too large (\eg, 0.9999), due to over-smoothing, it will lead to under-fitting in insufficient training schedule; When $m$ is too small (\eg, 0 and 0.9), the performance drops considerably. Thus, the momentum should conform to the learning dynamics.

\subsection{Benchmarks}
We comprehensively compare our proposed method with the recently leading approaches in the two domain adaptation scenarios: GTA5 $\rightarrow$ Cityscapes in Table~\ref{tab:gta5_to_city} and SYNTHIA $\rightarrow$ Cityscapes in Table~\ref{tab:synthia_to_city}. 
From Table~\ref{tab:gta5_to_city}, we have the following observations:  
Our TransDA-S arrives at the best mIoU score as 60.6, outperforming all existing methods by a large margin. Even looking at the adaptation gain, our TransDA achieves +21.4(=60.6-39.2) which is still better than +20.7(=57.5-36.8) by the state-of-the-art method ProDA~\cite{zhang2021prototypical}.  
From Table~\ref{tab:synthia_to_city}, we observe that naively replacing the backbone with Swin-S even suffers a small small performance degradation -0.9(=37.7-38.6). It implies that generalization is instead compromised by naively replacing the backbone.
On the other hand, our TransDA still consistently surpasses all compared methods.
Specially, looking at the adaptation gain, our TransDA achieves +24.5(=62.2-37.7) with a larger margin than +23.4(=62.0-38.6) by the ProDA~\cite{zhang2021prototypical},  which can be viewed as significant considering that the better performance is harder to optimize. Such adaption gain verifies the effectiveness of our proposed method.

\section{Conclusion}
In this paper, we present TransDA, a novel domain adaptive semantic segmentation based on local ViTs. 
Our contribution is the discovery of the severe high-frequency component problem of local ViTs on pseudo-label generation and feature alignment for target domain. To tackle the above issues, we propose the momentum network and dynamic of discrepancy measurement, to smooth the learning dynamics for target domain. 
Compared to these long-explored methods on conventional backbones, TransDA can unleash the great potential of traditional methods without incurring substantial technical complexity, leading to a new SOTA baseline based on ViTs in domain adaptive semantic segmentation.

{\small
\bibliographystyle{ieee_fullname}
\bibliography{egbib}

\begin{thebibliography}{10}\itemsep=-1pt

\bibitem{araslanov2021self}
Nikita Araslanov and Stefan Roth.
\newblock Self-supervised augmentation consistency for adapting semantic
  segmentation.
\newblock In {\em Proceedings of the IEEE/CVF Conference on Computer Vision and
  Pattern Recognition}, pages 15384--15394, 2021.

\bibitem{brown2020language}
Tom Brown, Benjamin Mann, Nick Ryder, Melanie Subbiah, Jared~D Kaplan, Prafulla
  Dhariwal, Arvind Neelakantan, Pranav Shyam, Girish Sastry, Amanda Askell,
  et~al.
\newblock Language models are few-shot learners.
\newblock {\em Advances in neural information processing systems},
  33:1877--1901, 2020.

\bibitem{PFAN_2019_CVPR}
Chaoqi Chen, Weiping Xie, Wenbing Huang, Yu Rong, Xinghao Ding, Yue Huang,
  Tingyang Xu, and Junzhou Huang.
\newblock Progressive feature alignment for unsupervised domain adaptation.
\newblock In {\em The IEEE Conference on Computer Vision and Pattern
  Recognition (CVPR)}, 2019.

\bibitem{chen2017deeplab}
Liang-Chieh Chen, George Papandreou, Iasonas Kokkinos, Kevin Murphy, and Alan~L
  Yuille.
\newblock Deeplab: Semantic image segmentation with deep convolutional nets,
  atrous convolution, and fully connected crfs.
\newblock {\em IEEE transactions on pattern analysis and machine intelligence},
  40(4):834--848, 2017.

\bibitem{chen2011co}
Minmin Chen, Kilian~Q Weinberger, and John Blitzer.
\newblock Co-training for domain adaptation.
\newblock {\em Advances in neural information processing systems}, 24, 2011.

\bibitem{chen2017no}
Yi-Hsin Chen, Wei-Yu Chen, Yu-Ting Chen, Bo-Cheng Tsai, Yu-Chiang Frank~Wang,
  and Min Sun.
\newblock No more discrimination: Cross city adaptation of road scene
  segmenters.
\newblock In {\em Proceedings of the IEEE International Conference on Computer
  Vision}, pages 1992--2001, 2017.

\bibitem{choi2019self}
Jaehoon Choi, Taekyung Kim, and Changick Kim.
\newblock Self-ensembling with gan-based data augmentation for domain
  adaptation in semantic segmentation.
\newblock In {\em Proceedings of the IEEE/CVF International Conference on
  Computer Vision}, pages 6830--6840, 2019.

\bibitem{chu2021twins}
Xiangxiang Chu, Zhi Tian, Yuqing Wang, Bo Zhang, Haibing Ren, Xiaolin Wei,
  Huaxia Xia, and Chunhua Shen.
\newblock Twins: Revisiting the design of spatial attention in vision
  transformers.
\newblock In {\em NeurIPS 2021}, 2021.

\bibitem{cordts2016cityscapes}
Marius Cordts, Mohamed Omran, Sebastian Ramos, Timo Rehfeld, Markus Enzweiler,
  Rodrigo Benenson, Uwe Franke, Stefan Roth, and Bernt Schiele.
\newblock The cityscapes dataset for semantic urban scene understanding.
\newblock In {\em Proceedings of the IEEE conference on computer vision and
  pattern recognition}, pages 3213--3223, 2016.

\bibitem{dosovitskiy2020image}
Alexey Dosovitskiy, Lucas Beyer, Alexander Kolesnikov, Dirk Weissenborn,
  Xiaohua Zhai, Thomas Unterthiner, Mostafa Dehghani, Matthias Minderer, Georg
  Heigold, Sylvain Gelly, et~al.
\newblock An image is worth 16x16 words: Transformers for image recognition at
  scale.
\newblock In {\em International Conference on Learning Representations}, 2020.

\bibitem{everingham2015pascal}
Mark Everingham, SM~Ali Eslami, Luc Van~Gool, Christopher~KI Williams, John
  Winn, and Andrew Zisserman.
\newblock The pascal visual object classes challenge: A retrospective.
\newblock {\em International journal of computer vision}, 111(1):98--136, 2015.

\bibitem{ganin2016domain}
Yaroslav Ganin, Evgeniya Ustinova, Hana Ajakan, Pascal Germain, Hugo
  Larochelle, Fran{\c{c}}ois Laviolette, Mario Marchand, and Victor Lempitsky.
\newblock Domain-adversarial training of neural networks.
\newblock {\em The journal of machine learning research}, 17(1):2096--2030,
  2016.

\bibitem{goodfellow2014generative}
Ian Goodfellow, Jean Pouget-Abadie, Mehdi Mirza, Bing Xu, David Warde-Farley,
  Sherjil Ozair, Aaron Courville, and Yoshua Bengio.
\newblock Generative adversarial nets.
\newblock {\em Advances in neural information processing systems}, 27, 2014.

\bibitem{guo2021sotr}
Ruohao Guo, Dantong Niu, Liao Qu, and Zhenbo Li.
\newblock Sotr: Segmenting objects with transformers.
\newblock In {\em Proceedings of the IEEE/CVF International Conference on
  Computer Vision}, pages 7157--7166, 2021.

\bibitem{guo2021metacorrection}
Xiaoqing Guo, Chen Yang, Baopu Li, and Yixuan Yuan.
\newblock Metacorrection: Domain-aware meta loss correction for unsupervised
  domain adaptation in semantic segmentation.
\newblock In {\em Proceedings of the IEEE/CVF Conference on Computer Vision and
  Pattern Recognition}, pages 3927--3936, 2021.

\bibitem{han2021demystifying}
Qi Han, Zejia Fan, Qi Dai, Lei Sun, Ming-Ming Cheng, Jiaying Liu, and Jingdong
  Wang.
\newblock Demystifying local vision transformer: Sparse connectivity, weight
  sharing, and dynamic weight.
\newblock {\em arXiv preprint arXiv:2106.04263}, 2021.

\bibitem{han2021dynamic}
Yizeng Han, Gao Huang, Shiji Song, Le Yang, Honghui Wang, and Yulin Wang.
\newblock Dynamic neural networks: A survey.
\newblock {\em arXiv preprint arXiv:2102.04906}, 2021.

\bibitem{he2016deep}
Kaiming He, Xiangyu Zhang, Shaoqing Ren, and Jian Sun.
\newblock Deep residual learning for image recognition.
\newblock In {\em Proceedings of the IEEE conference on computer vision and
  pattern recognition}, pages 770--778, 2016.

\bibitem{hinton2015distilling}
Geoffrey Hinton, Oriol Vinyals, Jeff Dean, et~al.
\newblock Distilling the knowledge in a neural network.
\newblock {\em arXiv preprint arXiv:1503.02531}, 2(7), 2015.

\bibitem{hoffman2018cycada}
Judy Hoffman, Eric Tzeng, Taesung Park, Jun-Yan Zhu, Phillip Isola, Kate
  Saenko, Alexei Efros, and Trevor Darrell.
\newblock Cycada: Cycle-consistent adversarial domain adaptation.
\newblock In {\em International conference on machine learning}, pages
  1989--1998. PMLR, 2018.

\bibitem{hoffman2016fcns}
Judy Hoffman, Dequan Wang, Fisher Yu, and Trevor Darrell.
\newblock Fcns in the wild: Pixel-level adversarial and constraint-based
  adaptation, 2016.

\bibitem{kenton2019bert}
Jacob Devlin Ming-Wei~Chang Kenton and Lee~Kristina Toutanova.
\newblock Bert: Pre-training of deep bidirectional transformers for language
  understanding.
\newblock In {\em Proceedings of NAACL-HLT}, pages 4171--4186, 2019.

\bibitem{kim2020learning}
Myeongjin Kim and Hyeran Byun.
\newblock Learning texture invariant representation for domain adaptation of
  semantic segmentation.
\newblock In {\em Proceedings of the IEEE/CVF conference on computer vision and
  pattern recognition}, pages 12975--12984, 2020.

\bibitem{kingma2014adam}
Diederik~P Kingma and Jimmy Ba.
\newblock Adam: A method for stochastic optimization.
\newblock {\em arXiv preprint arXiv:1412.6980}, 2014.

\bibitem{lee2013pseudo}
Dong-Hyun Lee et~al.
\newblock Pseudo-label: The simple and efficient semi-supervised learning
  method for deep neural networks.
\newblock In {\em Workshop on challenges in representation learning, ICML},
  volume~3, page 896, 2013.

\bibitem{li2021dynamic}
Yunsheng Li, Lu Yuan, Yinpeng Chen, Pei Wang, and Nuno Vasconcelos.
\newblock Dynamic transfer for multi-source domain adaptation.
\newblock In {\em Proceedings of the IEEE/CVF Conference on Computer Vision and
  Pattern Recognition}, pages 10998--11007, 2021.

\bibitem{li2019bidirectional}
Yunsheng Li, Lu Yuan, and Nuno Vasconcelos.
\newblock Bidirectional learning for domain adaptation of semantic
  segmentation.
\newblock In {\em Proceedings of the IEEE Conference on Computer Vision and
  Pattern Recognition}, pages 6936--6945, 2019.

\bibitem{liu2021Swin}
Ze Liu, Yutong Lin, Yue Cao, Han Hu, Yixuan Wei, Zheng Zhang, Stephen Lin, and
  Baining Guo.
\newblock Swin transformer: Hierarchical vision transformer using shifted
  windows.
\newblock {\em International Conference on Computer Vision (ICCV)}, 2021.

\bibitem{long2015fully}
Jonathan Long, Evan Shelhamer, and Trevor Darrell.
\newblock Fully convolutional networks for semantic segmentation.
\newblock In {\em Proceedings of the IEEE conference on computer vision and
  pattern recognition}, pages 3431--3440, 2015.

\bibitem{loshchilov2018decoupled}
Ilya Loshchilov and Frank Hutter.
\newblock Decoupled weight decay regularization.
\newblock In {\em International Conference on Learning Representations}, 2018.

\bibitem{luo2019taking}
Yawei Luo, Liang Zheng, Tao Guan, Junqing Yu, and Yi Yang.
\newblock Taking a closer look at domain shift: Category-level adversaries for
  semantics consistent domain adaptation.
\newblock In {\em Proceedings of the IEEE Conference on Computer Vision and
  Pattern Recognition}, pages 2507--2516, 2019.

\bibitem{ma2021coarse}
Haoyu Ma, Xiangru Lin, Zifeng Wu, and Yizhou Yu.
\newblock Coarse-to-fine domain adaptive semantic segmentation with photometric
  alignment and category-center regularization.
\newblock In {\em Proceedings of the IEEE/CVF Conference on Computer Vision and
  Pattern Recognition}, pages 4051--4060, 2021.

\bibitem{mcinnes2018umap}
Leland McInnes, John Healy, and James Melville.
\newblock Umap: Uniform manifold approximation and projection for dimension
  reduction.
\newblock {\em arXiv preprint arXiv:1802.03426}, 2018.

\bibitem{mei2020instance}
Ke Mei, Chuang Zhu, Jiaqi Zou, and Shanghang Zhang.
\newblock Instance adaptive self-training for unsupervised domain adaptation.
\newblock In {\em European conference on computer vision}, pages 415--430.
  Springer, 2020.

\bibitem{pan2020unsupervised}
Fei Pan, Inkyu Shin, Francois Rameau, Seokju Lee, and In~So Kweon.
\newblock Unsupervised intra-domain adaptation for semantic segmentation
  through self-supervision.
\newblock In {\em Proceedings of the IEEE/CVF Conference on Computer Vision and
  Pattern Recognition}, pages 3764--3773, 2020.

\bibitem{ren2015faster}
Shaoqing Ren, Kaiming He, Ross Girshick, and Jian Sun.
\newblock Faster r-cnn: Towards real-time object detection with region proposal
  networks.
\newblock {\em Advances in neural information processing systems}, 28, 2015.

\bibitem{ren2021better}
Yi Ren, Shangmin Guo, and Danica~J Sutherland.
\newblock Better supervisory signals by observing learning paths.
\newblock In {\em International Conference on Learning Representations}, 2021.

\bibitem{richter2016playing}
Stephan~R Richter, Vibhav Vineet, Stefan Roth, and Vladlen Koltun.
\newblock Playing for data: Ground truth from computer games.
\newblock In {\em European conference on computer vision}, pages 102--118.
  Springer, 2016.

\bibitem{ros2016synthia}
German Ros, Laura Sellart, Joanna Materzynska, David Vazquez, and Antonio~M
  Lopez.
\newblock The synthia dataset: A large collection of synthetic images for
  semantic segmentation of urban scenes.
\newblock In {\em Proceedings of the IEEE conference on computer vision and
  pattern recognition}, pages 3234--3243, 2016.

\bibitem{shimodaira2000improving}
Hidetoshi Shimodaira.
\newblock Improving predictive inference under covariate shift by weighting the
  log-likelihood function.
\newblock {\em Journal of statistical planning and inference}, 90(2):227--244,
  2000.

\bibitem{sun2019not}
Ruoqi Sun, Xinge Zhu, Chongruo Wu, Chen Huang, Jianping Shi, and Lizhuang Ma.
\newblock Not all areas are equal: Transfer learning for semantic segmentation
  via hierarchical region selection.
\newblock In {\em Proceedings of the IEEE/CVF Conference on Computer Vision and
  Pattern Recognition}, pages 4360--4369, 2019.

\bibitem{touvron2021training}
Hugo Touvron, Matthieu Cord, Matthijs Douze, Francisco Massa, Alexandre
  Sablayrolles, and Herv{\'e} J{\'e}gou.
\newblock Training data-efficient image transformers \& distillation through
  attention.
\newblock In {\em International Conference on Machine Learning}, pages
  10347--10357. PMLR, 2021.

\bibitem{tsai2018learning}
Yi-Hsuan Tsai, Wei-Chih Hung, Samuel Schulter, Kihyuk Sohn, Ming-Hsuan Yang,
  and Manmohan Chandraker.
\newblock Learning to adapt structured output space for semantic segmentation.
\newblock In {\em Proceedings of the IEEE Conference on Computer Vision and
  Pattern Recognition}, pages 7472--7481, 2018.

\bibitem{tzeng2017adversarial}
Eric Tzeng, Judy Hoffman, Kate Saenko, and Trevor Darrell.
\newblock Adversarial discriminative domain adaptation.
\newblock In {\em Proceedings of the IEEE conference on computer vision and
  pattern recognition}, pages 7167--7176, 2017.

\bibitem{vaswani2021scaling}
Ashish Vaswani, Prajit Ramachandran, Aravind Srinivas, Niki Parmar, Blake
  Hechtman, and Jonathon Shlens.
\newblock Scaling local self-attention for parameter efficient visual
  backbones, 2021.

\bibitem{vaswani2017attention}
Ashish Vaswani, Noam Shazeer, Niki Parmar, Jakob Uszkoreit, Llion Jones,
  Aidan~N Gomez, {\L}ukasz Kaiser, and Illia Polosukhin.
\newblock Attention is all you need.
\newblock In {\em Advances in neural information processing systems}, pages
  5998--6008, 2017.

\bibitem{wang2020classes}
Haoran Wang, Tong Shen, Wei Zhang, Ling-Yu Duan, and Tao Mei.
\newblock Classes matter: A fine-grained adversarial approach to cross-domain
  semantic segmentation.
\newblock In {\em European Conference on Computer Vision}, pages 642--659.
  Springer, 2020.

\bibitem{wang2021pyramid}
Wenhai Wang, Enze Xie, Xiang Li, Deng-Ping Fan, Kaitao Song, Ding Liang, Tong
  Lu, Ping Luo, and Ling Shao.
\newblock Pyramid vision transformer: A versatile backbone for dense prediction
  without convolutions.
\newblock In {\em Proceedings of the IEEE/CVF International Conference on
  Computer Vision}, pages 568--578, 2021.

\bibitem{xiao2018unified}
Tete Xiao, Yingcheng Liu, Bolei Zhou, Yuning Jiang, and Jian Sun.
\newblock Unified perceptual parsing for scene understanding.
\newblock In {\em Proceedings of the European Conference on Computer Vision
  (ECCV)}, pages 418--434, 2018.

\bibitem{xie2021segformer}
Enze Xie, Wenhai Wang, Zhiding Yu, Anima Anandkumar, Jose~M Alvarez, and Ping
  Luo.
\newblock Segformer: Simple and efficient design for semantic segmentation with
  transformers.
\newblock {\em arXiv preprint arXiv:2105.15203}, 2021.

\bibitem{yang2020fda}
Yanchao Yang and Stefano Soatto.
\newblock Fda: Fourier domain adaptation for semantic segmentation.
\newblock In {\em Proceedings of the IEEE/CVF Conference on Computer Vision and
  Pattern Recognition}, pages 4085--4095, 2020.

\bibitem{you2019universal}
Kaichao You, Mingsheng Long, Zhangjie Cao, Jianmin Wang, and Michael~I Jordan.
\newblock Universal domain adaptation.
\newblock In {\em Proceedings of the IEEE/CVF conference on computer vision and
  pattern recognition}, pages 2720--2729, 2019.

\bibitem{yuan2021tokens}
Li Yuan, Yunpeng Chen, Tao Wang, Weihao Yu, Yujun Shi, Zi-Hang Jiang,
  Francis~EH Tay, Jiashi Feng, and Shuicheng Yan.
\newblock Tokens-to-token vit: Training vision transformers from scratch on
  imagenet.
\newblock In {\em Proceedings of the IEEE/CVF International Conference on
  Computer Vision}, pages 558--567, 2021.

\bibitem{zhang2021prototypical}
Pan Zhang, Bo Zhang, Ting Zhang, Dong Chen, Yong Wang, and Fang Wen.
\newblock Prototypical pseudo label denoising and target structure learning for
  domain adaptive semantic segmentation.
\newblock In {\em Proceedings of the IEEE/CVF Conference on Computer Vision and
  Pattern Recognition}, pages 12414--12424, 2021.

\bibitem{zhao2017pyramid}
Hengshuang Zhao, Jianping Shi, Xiaojuan Qi, Xiaogang Wang, and Jiaya Jia.
\newblock Pyramid scene parsing network.
\newblock In {\em Proceedings of the IEEE conference on computer vision and
  pattern recognition}, pages 2881--2890, 2017.

\bibitem{zheng2021rectifying}
Zhedong Zheng and Yi Yang.
\newblock Rectifying pseudo label learning via uncertainty estimation for
  domain adaptive semantic segmentation.
\newblock {\em International Journal of Computer Vision}, 129(4):1106--1120,
  2021.

\end{thebibliography}
}
\clearpage

\begin{figure*}[t!]
    \centering
        \begin{subfigure}[b]{0.3\textwidth}
           \centering
            \includegraphics[width=\textwidth]{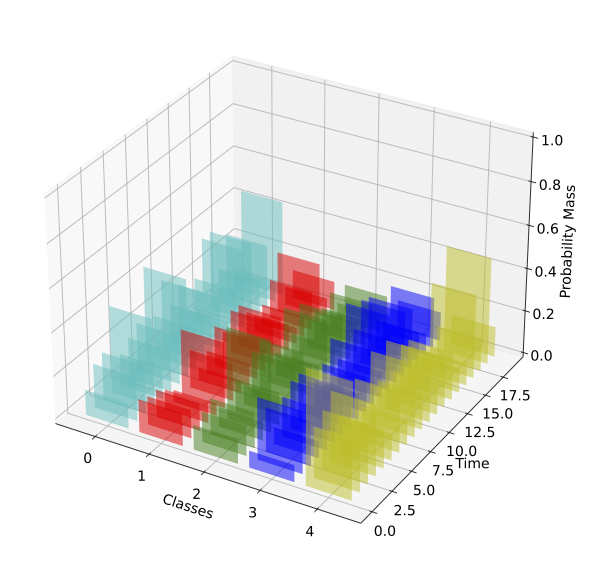}
            \caption{Predictions of `small' over time}
            \label{sfig:small_3d}
        \end{subfigure}
     \hfill
        \begin{subfigure}[b]{0.3\textwidth}
            \centering
            \includegraphics[width=\textwidth]{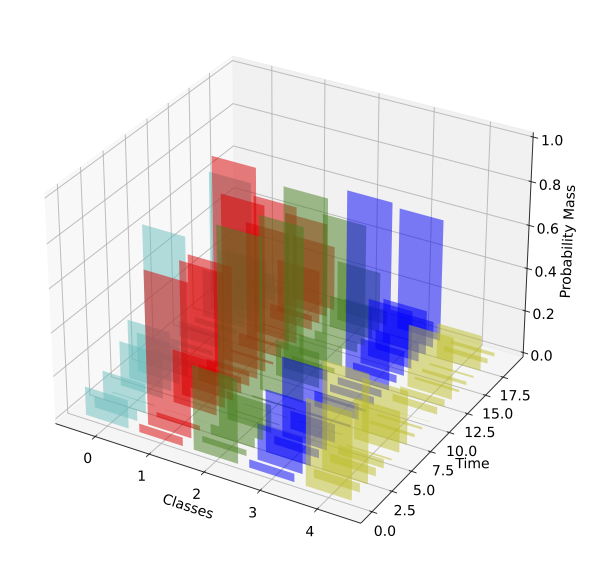}
            \caption{Predictions of `medium' over time}
            \label{sfig:medium_3d}
        \end{subfigure}
     \hfill
        \begin{subfigure}[b]{0.3\textwidth}
            \centering
            \includegraphics[width=\textwidth]{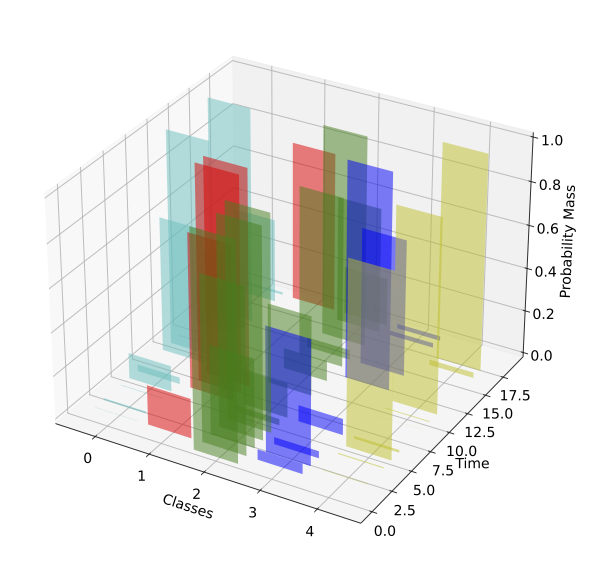}
            \caption{Predictions of `large' over time}
            \label{sfig:large_3d}
        \end{subfigure}
    \centering
        \begin{subfigure}[b]{0.3\textwidth}
           \centering
            \includegraphics[width=\textwidth]{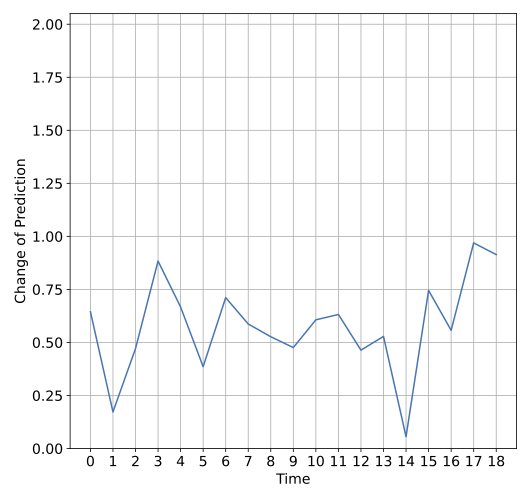}
            \caption{$\norm{\vp^s_{\ti+1} - \vp^s_{\ti}}_1$}
            \label{sfig:small_pred}
        \end{subfigure}
     \hfill
        \begin{subfigure}[b]{0.3\textwidth}
            \centering
            \includegraphics[width=\textwidth]{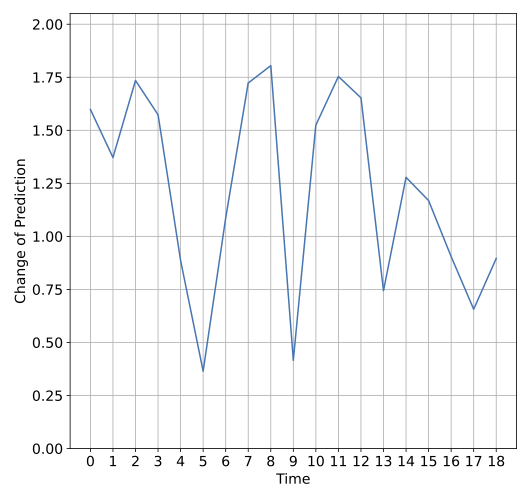}
            \caption{$\norm{\vp^m_{\ti+1} - \vp^m_{\ti}}_1$}
            \label{sfig:medium_pred}
        \end{subfigure}
     \hfill
        \begin{subfigure}[b]{0.3\textwidth}
            \centering
            \includegraphics[width=\textwidth]{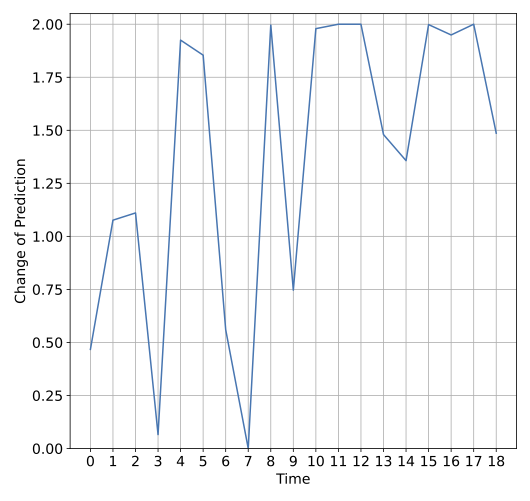}
            \caption{$\norm{\vp^l_{\ti+1} - \vp^l_{\ti}}_1$}
            \label{sfig:large_pred}
        \end{subfigure}
    \centering
        \begin{subfigure}[b]{0.3\textwidth}
           \centering
            \includegraphics[width=\textwidth]{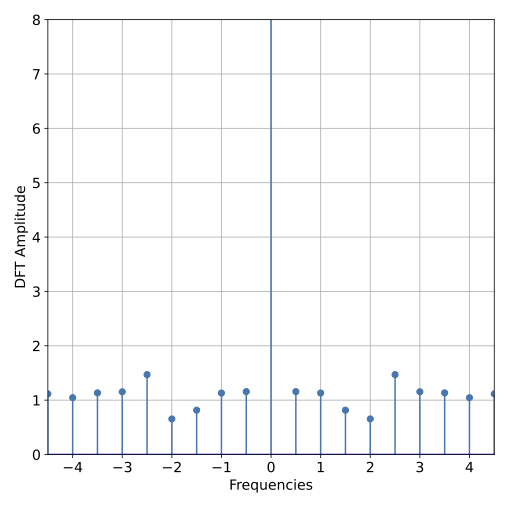}
            \caption{DFT of `small'}
            \label{sfig:small_dft}
        \end{subfigure}
     \hfill
        \begin{subfigure}[b]{0.3\textwidth}
            \centering
            \includegraphics[width=\textwidth]{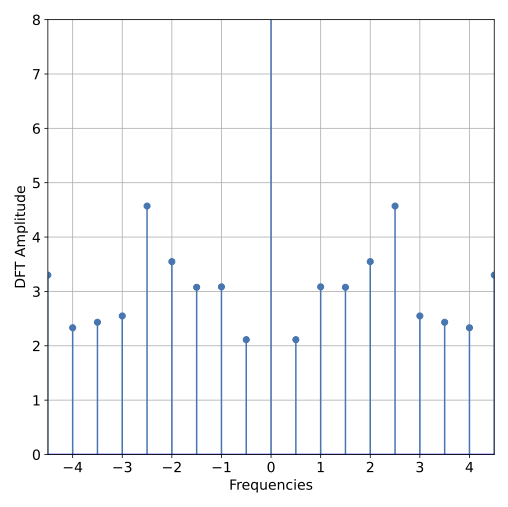}
            \caption{DFT of `medium'}
            \label{sfig:medium_dft}
        \end{subfigure}
     \hfill
        \begin{subfigure}[b]{0.3\textwidth}
            \centering
            \includegraphics[width=\textwidth]{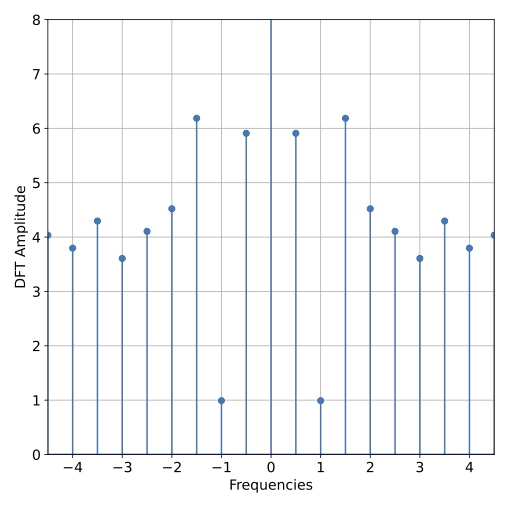}
            \caption{DFT of `large'}
            \label{sfig:large_dft}
        \end{subfigure}
    \caption{Illustration of High-Frequency Component Problem. 
    }
    \label{fig:high_freq_component}
\end{figure*}

\begin{appendix}
\section{Appendix}

\subsection{Toy Example to Illustrate High-Frequency Components Problem}

In Figure~1(b) and (c) of our main body, we track both $\norm{\mathbf{D}_{\ti+1}(\mathbf{F}^s_{\ti+1}(\vx^t)) - \mathbf{D}_{\ti}(\mathbf{F}^s_\ti(\vx^t))}_1$ and $\norm{\mathbf{C}^s_{\ti+1}(\mathbf{F}^s_{\ti+1}(\vx^t)) - \mathbf{C}^s_\ti(\mathbf{F}^s_\ti(\vx^t))}_1$. Denotes $\mathbf{D}_{\ti}(\mathbf{F}^s_{\ti}(\vx^t))$ or $\mathbf{C}^s_\ti(\mathbf{F}^s_\ti(\vx^t))$ as $\vp_\ti$. Then we refer to 
\begin{equation}
\footnotesize
\norm{\vp_{\ti+1} - \vp_{\ti}}_1 = \frac{1}{S*N}\sum_{s}^S\sum_{n}^N\sum_{w}^W\sum_{h}^H\sum_{k}^K \norm{\vp^{s,n,w,h,k}_{\ti+1}-\vp^{s,n,w,h,k}_\ti}
\label{eq.change}
\end{equation} as \emph{change of predictions} in local ViTs (Swin-S) for domain adaptive semantic segmentation, where $S=5$ denotes the number of seeds, $N=10$ denotes the number of images, $W$ and $H$ denotes the size of predictions, $K=19$ denotes the number of categories.

However, since the above $\vp_\ti$ is too complicated to visualize, in the following of this subsection, we instead use a toy example for visualisation and illustration about the so called high-frequency component problem which implies that \emph{large variance in model’s prediction} in this section.
Suppose the number of categories is $K=5$.
To control the variance degree of model's predictions over time, we synthesize the prediction sequences with three Gaussian processes whose means are $\bm{\mu}_1=\bm{\mu}_2=\bm{\mu}_3=\mathbf{0}\in\mathbb{R}^{5}$ and variances are $\sigma_1=0.5, \sigma_2=1.0, \sigma_3=10.0$.
To make the following easier, let's denote these three Gaussian processes as `small', `medium', `large' respectively.
We then sample predictions over $T=20$ iterations time for each of them, and then apply a $\softmax$ function along categorical dimension on these predictions to convert them to categorical distributions which are denoted as $\vp^s_\ti, \vp^m_\ti, \vp^l_\ti, \forall \ti \in T$ respectively.
The probability sequences of all categories are then plotted in Figure~\ref{sfig:small_3d},~\ref{sfig:medium_3d}, and~\ref{sfig:large_3d}.


Similar to Eq.~\ref{eq.change}, we refer to $\norm{\vp_{\ti+1} - \vp_{\ti}}_1 = \sum_{k}^K \norm{\vp^{k}_{\ti+1}-\vp^{k}_\ti}$
as \emph{change of predictions} in our toy example.
We plot $\norm{\vp^s_{\ti+1} - \vp^s_{\ti}}_1$, $\norm{\vp^m_{\ti+1} - \vp^m_{\ti}}_1$, and $\norm{\vp^l_{\ti+1} - \vp^l_{\ti}}_1$, $\forall \ti \in T$ in Figure~\ref{sfig:small_pred},~\ref{sfig:medium_pred}, and~\ref{sfig:large_pred}.
It is straightforward to see that the `large' process produce higher values since its predictions change more drastically.


We then can do a Discrete Fourier Transformation (DFT) on the series of $\vp^s_{t}$, $\vp^m_{\ti}$, and $\vp^l_{\ti}$, $\forall \ti \in T$, and the results are shown in Figure~\ref{sfig:small_dft},~\ref{sfig:medium_dft}, and~\ref{sfig:large_dft}.
From Figure~\ref{sfig:large_dft}, it can be straightforwardly seen that the amplitude of the high frequencies (\eg, 4 and -4 on the horizontal axis) are higher when the change of prediction are higher in Figure~\ref{sfig:large_pred}.
That said, the existence of more high-frequency components implies \emph{more drastic change of model's predictions}.
Therefore, we refer it as the ``high-frequency component'' problem.

\subsection{More Discussions about the High-Frequency Components Problem}

\paragraph{More Observations of the Problem.} 
In Supplementary Material, we further explore the high-frequency component problem on source domain.
We randomly sample 8 images from source domain, GTA5~\cite{richter2016playing} and target domain, Cityscapes train set~\cite{cordts2016cityscapes}, respectively and use $S=5$ random seeds to run the experiment.

\begin{figure*}[!t]
    \centering
    \includegraphics[width=0.9\linewidth]{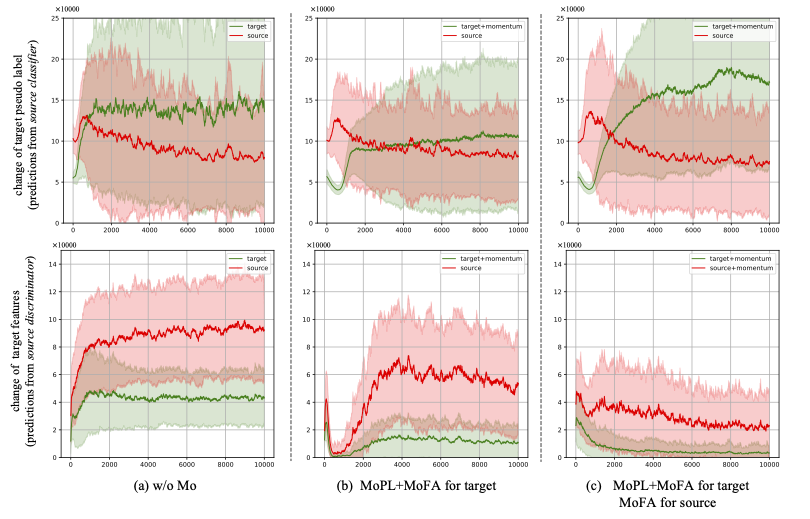}
    \vspace{-1.5em}
    \caption{High-frequency components problem in Swin-S ViT on source domain and target domain. 
    }
    \label{fig:swin}
\end{figure*}

First, we conduct research on Swin-S ViT and observe: 
\begin{itemize}[leftmargin=*]
\item In Figure~\ref{fig:swin}(a), without smoothing, the change of predictions $\mathbf{C}^s(\mathbf{F}^s(\vx^t))$  from classifier for target domain is large than $\mathbf{C}^s(\mathbf{F}^s(\vx^s))$ for source domain, but the opposite in $\mathbf{D}(\mathbf{F}^s(\vx^t))$ and $\mathbf{D}(\mathbf{F}^s(\vx^s))$ from discriminator. 
This may due to that in adversarial training, the features of target domain $\mathbf{F}^s(\vx^t)$  encoding by the source feature extractor are difficult to deceive the discriminator, but the features of source domain $\mathbf{F}^s(\vx^s)$  can easily deceive the discriminator during the training process.
\item In Figure~\ref{fig:swin}(b), after applying momentum networks $\mathbf{F}^t$ and $\mathbf{C}^t$ for target domain,  the change of predictions $\mathbf{C}^t(\mathbf{F}^t(\vx^t))$ from classifier for target domain is significantly reduced, and the change of predictions $\mathbf{D}(\mathbf{F}^s(\vx^s))$ and $\mathbf{D}(\mathbf{F}^t(\vx^t))$ from discriminator are also both pulled down, which might because the features of source domain is indirectly smoothed due to feature alignment.
\item For the high-frequency components problem on source domain, we only consider smoothing the source domain features for feature alignment, due to that source domain has ground-truth supervision signals. 
In Figure~\ref{fig:swin}(c), after applying momentum network $\mathbf{F}^t$  for source domain,  the change of predictions $\mathbf{D}(\mathbf{F}^t(\vx^s))$ and $\mathbf{D}(\mathbf{F}^t(\vx^t))$ from discriminator are both pulled down further. However, this caused the change of predictions $\mathbf{C}^t(\mathbf{F}^t(\vx^t))$ from classifier for target domain to oscillate violently.
The performance of this variant on the task of GTA5 $\rightarrow$ Cityscapes ($57.8_{\pm0.8}$)
 shows that smoothing the source domain features for feature alignment will cause performance degradation.
\end{itemize}

\begin{figure*}[!t]
    \centering
    \includegraphics[width=0.9\linewidth]{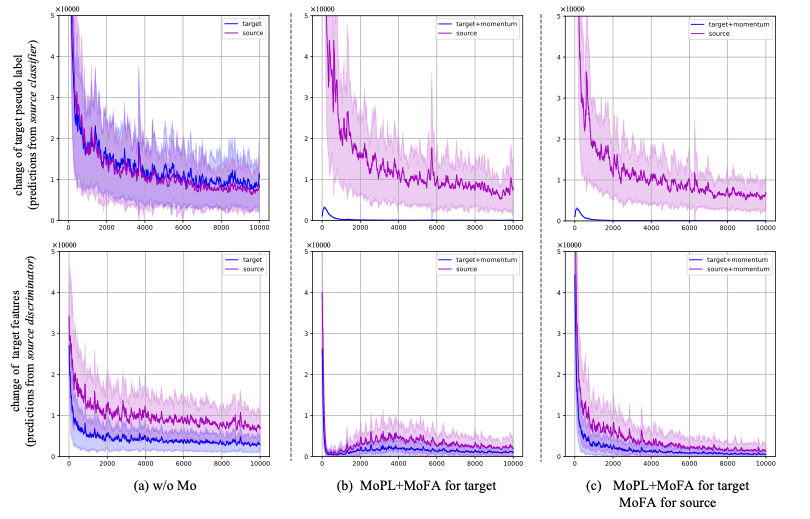}
    \vspace{-1.5em}
    \caption{High-frequency components problem in ResNet-101 on source domain and target domain. 
    }
    \label{fig:resnet}
\end{figure*}

Secondly, we also conduct research on ResNet-101, and observe: 
\begin{itemize}[leftmargin=*]
\item In contrast to Figure~\ref{fig:swin}, we can find the change of predictions is very small in Figure~\ref{fig:resnet}, which implies that the high-frequency component problem is specific for ViTs but not CNNs.
\item In Figure~\ref{fig:resnet}(a) and (b), the situation is basically the same as that of Figure~\ref{fig:swin}(a) and (b).
\item In Figure~\ref{fig:resnet}(c), smoothing the source domain features for feature alignment is basically useless. Compared with Figure~\ref{fig:swin}(c), the change of predictions $\mathbf{C}^t(\mathbf{F}^t( \vx^t))$ from classifier is also not affected. 
\end{itemize}

\paragraph{Impact of the Issue.}

The impact of high-frequency components in the predictions of the classifier is reflected in generating target pseudo labels after each training round, which is not conducive to \textbf{self-training in the next training round}.
In contrast, the impact of high-frequency components in feature encoding is mainly reflected in features alignment after each training iteration, which is not conducive to \textbf{adversarial training in the next training iteration}.
The period and phase of impact are different, target features affect each iteration in the \emph{current round}, but target pseudo labels affect each iteration in the \emph{next round}.
It is worth noting that adversarial training itself is unstable, let alone to align the poor target domain distribution produced by ViT, it will be even more unstable.
So smoothing could be a requirement for ViT with adversarial training in various tasks.
Different from \cite{ren2021better}, our work discusses the importance of features in adversarial training for domain adaptation, while \cite{ren2021better} emphasizes the importance of labels in supervised training.

\subsection{Details of Discrepancy Measurements\\ in Table 1}

In this section, we provide the detailed definitions about the discrepancy measurements $\text{Dis}(\cdot)$ that we compared in the ablation study: smoothing matters.

For binary domain discriminator, \emph{the binary adversarial loss} is defined as follows:
\begin{equation}
\aligned
\mathcal{L}^{bin}_{adv} &=  -\mathbb{E}_{\gD^s} [\log \mathbf{D}(\mathbf{F}^s(\vx^s)))]\\
                & -\mathbb{E}_{\gD^t} [\log (1-\mathbf{D}(\mathbf{F}^t(\vx^t)))].
\endaligned
\label{eq.bin}
\end{equation}
We realize the dynamic discrepancy $\text{Dis}_{w(\vx^s,\vx^t)}$ by \emph{the weighted binary adversarial loss} below:
\begin{equation}
\aligned
\mathcal{L}^{wbin}_{adv} &=  -\mathbb{E}_{\gD^s} [w^s(\vx^s)\log \mathbf{D}(\mathbf{F}^s(\vx^s)))]\\
                & -\mathbb{E}_{\gD^t} [w^t(\vx^t)\log (1-\mathbf{D}(\mathbf{F}^t(\vx^t)))],
\endaligned
\label{eq.wbin}
\end{equation}
where the dynamic weighting $w(\vx^s, \vx^t)$ refers to $w^s(\vx^s)$ on the source input and $w^t(\vx^t)$ on the target input.

For class domain discriminator, \emph{the class-level adversarial loss} is:
\begin{align}
\label{eq.cls}
\begin{split}
	\mathcal{L}^{cls}_{adv} &=  -\mathbb{E}_{\gD^s} [ \sum_{k=1}^{K^c} p^s_k \log \mathbf{D}_k(\mathbf{F}^s(\vx^s))] \\
    & -\mathbb{E}_{\gD^t}  [ \sum_{k=1}^{K^c} \hat{p}^t_k \log (1-\mathbf{D}_{k}(\mathbf{F}^t(\vx^t)))],
\end{split}
\end{align}
where $\mathbf{D}_k$ refers to the $k$-th output channel of the discriminator, and the prediction $p_k$ from the classifier is adopted to balance the class-level importance. 
We implement the dynamic discrepancy $\text{Dis}_{w(\vx^s,\vx^t)}$ as \emph{a weighted class-level adversarial loss} defined as below:
\begin{align}
\label{eq.wcls}
\begin{split}
	\mathcal{L}^{wcls}_{adv} &=  -\mathbb{E}_{\gD^s} [w^s(\vx^s) \sum_{k=1}^{K^c} p^s_k \log \mathbf{D}_k(\mathbf{F}^s(\vx^s))] \\
    & -\mathbb{E}_{\gD^t}  [w^t(\vx^t) \sum_{k=1}^{K^c} \hat{p}^t_k \log (1-\mathbf{D}_{k}(\mathbf{F}^t(\vx^t)))].
\end{split}
\end{align}

\subsection{Averaged Scores and Standard Deviations on Benchmarks}

We report the mean mIoU with standard deviations to demonstrate the robustness of TransDA in the two domain adaptation scenarios: GTA5 $\rightarrow$ Cityscapes in Table~\ref{tab:gta5_to_city_ave} and SYNTHIA $\rightarrow$ Cityscapes in Table~\ref{tab:synthia_to_city_ave}.  
The low standard deviations demonstrate the stability of our method. 
Even looking in the sense of averaged mIoU, our TransDA is still surpassing or matching the best score by the recently leading approaches in these two domain adaptation scenarios. 

Besides, we can find that the generalization performance can be significantly improved as the scale of model parameters increases (+5.8=44.1-38.4 in GTA5$\rightarrow$ Cityscapes and +7.5=43.7-36.2 in  SYNTHIA$\rightarrow$ Cityscapes). However, the gains of the adaptation are obviously limited (+2.5=61.8-59.3 in GTA5$\rightarrow$ Cityscapes and +3.9=65.4-61.5 in  SYNTHIA$\rightarrow$ Cityscapes), considering that the better performance is harder to optimize.

For discrepancy measurement, we find that adversarial training based on binary discriminator is more stable than class discriminator. %
Although the per-class distribution alignment is in a more fine-grained manner~\cite{wang2020classes,chen2017no},the problem is that such alignment heavily depends on the class supervision from the source and target domain. 
Therefore, as the scale of the model parameters increases, the class-level method based on the Swin-B backbone can better predict the self-supervised signal, resulting in comparable performance (see last row in Table~\ref{tab:gta5_to_city_ave} and Table~\ref{tab:synthia_to_city_ave}), which bodes well for \textbf{greater potential}.

\begin{table*}[!t]
\centering
\resizebox{\textwidth}{!}{
\begin{tabular}{c|*{19}{c}|c}
\toprule 
  & \multicolumn{1}{c}{\begin{turn}{60}road\end{turn}} & \multicolumn{1}{c}{\begin{turn}{60}sideway\end{turn}} & \multicolumn{1}{c}{\begin{turn}{60}building\end{turn}} & \multicolumn{1}{c}{\begin{turn}{60}wall\end{turn}} & \multicolumn{1}{c}{\begin{turn}{60}fence\end{turn}} & \multicolumn{1}{c}{\begin{turn}{60}pole\end{turn}} & \multicolumn{1}{c}{\begin{turn}{60}light\end{turn}} & \multicolumn{1}{c}{\begin{turn}{60}sign\end{turn}} & \multicolumn{1}{c}{\begin{turn}{60}vege.\end{turn}} & \multicolumn{1}{c}{\begin{turn}{60}terrace\end{turn}} & \multicolumn{1}{c}{\begin{turn}{60}sky\end{turn}} & \multicolumn{1}{c}{\begin{turn}{60}person\end{turn}} & \multicolumn{1}{c}{\begin{turn}{60}rider\end{turn}} & \multicolumn{1}{c}{\begin{turn}{60}car\end{turn}} & \multicolumn{1}{c}{\begin{turn}{60}truck\end{turn}} & \multicolumn{1}{c}{\begin{turn}{60}bus\end{turn}} & \multicolumn{1}{c}{\begin{turn}{60}train\end{turn}} & \multicolumn{1}{c}{\begin{turn}{60}motor\end{turn}} & \multicolumn{1}{c}{\begin{turn}{60}bike\end{turn}} & \multicolumn{1}{|c}{mIoU}\\
\midrule 
\multicolumn{21}{l}{\emph{Backbone: Swin-S ViT (50M)}}\\
\midrule 
{w/o Adaptation}  &51.4\tiny3.3&22.8\tiny3.8&55.8\tiny5.2&14.5\tiny2.1&17.1\tiny3.5&26.8\tiny3.7&42.8\tiny3.7&19.6\tiny1.6&83.6\tiny0.7&32.4\tiny1.5&80.9\tiny6.1&63.2\tiny0.9&28.1\tiny2.5&66.9\tiny16.3&25.4\tiny7.6&26.2\tiny10.6&3.1\tiny3.1&32.9\tiny1.9&33.4\tiny5.3&38.3\tiny0.8\\
{TransDA-S} &90.3\tiny2.9&55.8\tiny4.1&85.9\tiny2.3&41.5\tiny1.7&26.8\tiny2.9&46.8\tiny1.1&57.2\tiny0.2&40.1\tiny1.6&90.1\tiny0.3&46.3\tiny2.5&93.8\tiny0.2&74.3\tiny0.3&44.2\tiny1.2&91.4\tiny0.2&49.5\tiny2.9&56.3\tiny4.2&38.6\tiny10.6&44.6\tiny4.8&52.4\tiny0.9&59.3\tiny0.7\\
\midrule 
\multicolumn{21}{l}{\emph{Backbone: Swin-B ViT (88M)}}\\
\midrule 
{w/o Adaptation} &72.6\tiny7.0&23.6\tiny3.7&72.3\tiny3.5&16.9\tiny3.3&24.1\tiny2.5&36.6\tiny4.2&49.4\tiny1.2&30.6\tiny4.8&82.9\tiny1.5&34.1\tiny5.7&84.4\tiny2.8&68.6\tiny1.4&29.1\tiny2.8&74.7\tiny9.3&31.1\tiny3.7&20.0\tiny6.1&8.8\tiny5.9&37.6\tiny2.0&40.1\tiny1.6&44.1\tiny0.5\\
{TransDA-B} &91.7\tiny0.8&55.8\tiny2.0&89.1\tiny0.2&46.7\tiny0.7&39.2\tiny4.9&49.8\tiny0.7&60.3\tiny0.3&44.5\tiny1.1&90.7\tiny0.1&51.4\tiny0.5&93.9\tiny0.1&75.2\tiny0.2&45.5\tiny0.2&92.4\tiny0.2&50.8\tiny3.8&51.4\tiny5.6&48.3\tiny3.2&43.5\tiny7.6&54.1\tiny3.6&61.8\tiny0.9\\
{TransDA-B$^{wcls}$} &92.7\tiny1.1&58.5\tiny4.0&89.2\tiny0.2&48.9\tiny1.1&42.4\tiny3.4&50.1\tiny0.4&60.3\tiny0.3&43.9\tiny1.8&90.5\tiny0.2&50.7\tiny1.0&93.7\tiny0.2&75.7\tiny0.7&46.6\tiny0.9&92.1\tiny0.3&53.5\tiny3.3&57.4\tiny4.2&29.3\tiny11.9&45.6\tiny4.1&55.2\tiny2.6&61.9\tiny1.2\\

\bottomrule  
\end{tabular}%
}
\vspace{-.5em}
\caption { The averaged scores and standard deviations  on GTA5$\to$Cityscapes. \emph{wcls} denotes employing the weighted class-level adversarial loss in Eq.~\ref{eq.wcls}}
\label{tab:gta5_to_city_ave}%
\end{table*}%

\begin{table*}[!t]
\centering
\resizebox{\textwidth}{!}{
\begin{tabular}{@{}c|*{16}{c}|c|c@{}}
\toprule
 & \multicolumn{1}{c}{\begin{turn}{60}road\end{turn}} & \multicolumn{1}{c}{\begin{turn}{60}sideway\end{turn}} & \multicolumn{1}{c}{\begin{turn}{60}building\end{turn}} & \multicolumn{1}{c}{\begin{turn}{60}wall*\end{turn}} & \multicolumn{1}{c}{\begin{turn}{60}fence*\end{turn}} & \multicolumn{1}{c}{\begin{turn}{60}pole*\end{turn}} & \multicolumn{1}{c}{\begin{turn}{60}light\end{turn}} & \multicolumn{1}{c}{\begin{turn}{60}sign\end{turn}} & \multicolumn{1}{c}{\begin{turn}{60}vege.\end{turn}} & \multicolumn{1}{c}{\begin{turn}{60}sky\end{turn}} & \multicolumn{1}{c}{\begin{turn}{60}person\end{turn}} & \multicolumn{1}{c}{\begin{turn}{60}rider\end{turn}} & \multicolumn{1}{c}{\begin{turn}{60}car\end{turn}}& \multicolumn{1}{c}{\begin{turn}{60}bus\end{turn}} & \multicolumn{1}{c}{\begin{turn}{60}motor\end{turn}} & \multicolumn{1}{c}{\begin{turn}{60}bike\end{turn}}& \multicolumn{1}{|l}{mIoU} & \multicolumn{1}{|l}{mIoU*}\\
\midrule 
\multicolumn{19}{l}{\emph{Backbone: Swin-S ViT (50M)}}\\
\midrule 

{w/o Adaptation}  &25.2\tiny4.5&24.5\tiny2.2&39.6\tiny2.4&4.0\tiny0.4&0.1\tiny0.0&26.5\tiny1.6&32.4\tiny0.7&15.6\tiny0.1&79.9\tiny3.1&67.9\tiny2.7&55.5\tiny3.6&6.1\tiny1.5&73.8\tiny3.5&30.0\tiny2.4&9.5\tiny1.3&11.2\tiny3.0&31.4\tiny0.8&36.2\tiny1.0\\
{TransDA-S} &82.1\tiny1.8&43.6\tiny2.3&83.2\tiny2.7&18.9\tiny4.5&1.1\tiny0.4&49.3\tiny2.4&56.4\tiny1.9&29.9\tiny3.8&89.9\tiny0.5&92.5\tiny1.4&65.8\tiny1.7&20.4\tiny3.5&91.0\tiny0.2&57.2\tiny5.5&41.0\tiny0.7&48.3\tiny2.4&54.4\tiny0.7&61.5\tiny0.4\\
\midrule 
\multicolumn{19}{l}{\emph{Backbone: Swin-B ViT (88M)}}\\
\midrule 
{w/o Adaptation} &43.6\tiny10.7&30.3\tiny4.6&56.4\tiny2.9&11.1\tiny4.6&0.3\tiny0.1&36.8\tiny2.3&34.3\tiny3.3&20.8\tiny2.4&82.2\tiny1.9&78.2\tiny3.0&63.6\tiny0.7&13.1\tiny1.5&67.7\tiny15.8&36.2\tiny3.6&20.5\tiny2.3&21.0\tiny1.3&38.5\tiny0.5&43.7\tiny1.1\\
{TransDA-B} &83.7\tiny2.5&43.6\tiny3.2&86.5\tiny0.6&27.3\tiny3.2&1.2\tiny0.3&54.3\tiny0.7&61.0\tiny0.3&36.4\tiny1.9&90.1\tiny0.5&92.4\tiny1.7&71.1\tiny0.6&27.4\tiny2.9&92.4\tiny0.3&66.4\tiny4.0&49.6\tiny1.6&50.1\tiny0.6&58.3\tiny0.5&65.4\tiny0.5\\
{TransDA-B$^{wcls}$} &84.6\tiny3.3&46.0\tiny5.1&85.8\tiny1.3&25.4\tiny4.1&1.3\tiny0.3&54.6\tiny0.6&61.1\tiny0.5&38.1\tiny4.4&90.4\tiny0.2&93.5\tiny0.5&71.1\tiny0.9&26.2\tiny1.4&91.1\tiny2.0&64.9\tiny3.1&47.6\tiny2.2&50.1\tiny1.2&58.3\tiny0.6&65.4\tiny0.6\\
\bottomrule  
\end{tabular}%
}
\vspace{-.5em}
\caption {The averaged scores and standard deviations  on SYNTHIA$\to$Cityscapes.  
  mIoU and mIoU* denote the scores across 16 and 13 categories respectively. 
  \emph{wcls} denotes employing the weighted class-level adversarial loss in Eq.~\ref{eq.wcls}}
\label{tab:synthia_to_city_ave}%
\end{table*}%

\begin{table*}[!t]
\centering
\resizebox{\textwidth}{!}{
\begin{tabular}{cc|*{19}{c}|c|c}
\toprule 
 & & \multicolumn{1}{c}{\begin{turn}{60}road\end{turn}} & \multicolumn{1}{c}{\begin{turn}{60}sideway\end{turn}} & \multicolumn{1}{c}{\begin{turn}{60}building\end{turn}} & \multicolumn{1}{c}{\begin{turn}{60}wall*\end{turn}} & \multicolumn{1}{c}{\begin{turn}{60}fence*\end{turn}} & \multicolumn{1}{c}{\begin{turn}{60}pole*\end{turn}} & \multicolumn{1}{c}{\begin{turn}{60}light\end{turn}} & \multicolumn{1}{c}{\begin{turn}{60}sign\end{turn}} & \multicolumn{1}{c}{\begin{turn}{60}vege.\end{turn}} & \multicolumn{1}{c}{\begin{turn}{60}terrace\end{turn}} & \multicolumn{1}{c}{\begin{turn}{60}sky\end{turn}} & \multicolumn{1}{c}{\begin{turn}{60}person\end{turn}} & \multicolumn{1}{c}{\begin{turn}{60}rider\end{turn}} & \multicolumn{1}{c}{\begin{turn}{60}car\end{turn}} & \multicolumn{1}{c}{\begin{turn}{60}truck\end{turn}} & \multicolumn{1}{c}{\begin{turn}{60}bus\end{turn}} & \multicolumn{1}{c}{\begin{turn}{60}train\end{turn}} & \multicolumn{1}{c}{\begin{turn}{60}motor\end{turn}} & \multicolumn{1}{c}{\begin{turn}{60}bike\end{turn}} & \multicolumn{1}{|c}{mIoU} & \multicolumn{1}{|l}{mIoU*}\\
\midrule
\multicolumn{23}{l}{\textit{GTA5$\rightarrow$Cityscapes}}\\
\midrule
\multicolumn{2}{c|}{SAC~\cite{araslanov2021self}}& 91.8 & 54.3 & 87.4 & 36.2 & 30.2 & 43.7 & 49.7 & 42.1 & 89.3 & 54.3 & 90.5 & 71.8 & 34.9 & 89.8 & 38.8 & 47.3 & 24.9 & 38.3 & 43.8 & 55.7 & - \\
\multicolumn{2}{c|}{TransDA-S}  &94.5&61.1&88.8&40.7&33.9&46.8&56.9&37.2&91.3&56.4&94.8&78.6&55.0&92.1&44.2&44.4&43.9&54.7&47.0  & {61.2} & - \\
\multicolumn{2}{c|}{TransDA-B}  & 95.4&63.0&89.7&45.8&44.2&49.2&59.4&38.0&91.7&60.0&94.5&80.8&57.4&93.2&43.5&47.3&38.7&57.2&50.9&63.2& - \\
 	 	 	 	 	 	 	 		 	 	 	 	 	 	 	 		
\midrule
\multicolumn{23}{l}{\textit{SYNTHIA$\rightarrow$Cityscapes}}\\
\midrule
\multicolumn{2}{c|}{SAC~\cite{araslanov2021self}}& 87.4 & 41.0 & 85.5 & 17.5 & 2.6 & 40.5 & 44.7 & 34.4 & 87.9 & - & 91.2 & 68.0 & 31.0 & 89.3 & - & 33.2 & - & 38.6 & 49.9 & 52.7 & 60.2\\
\multicolumn{2}{c|}{TransDA-S}  &76.6&30.6&84.8&11.4&1.5&48.7&53.2&37.5&87.4&-&91.9&70.8&29.8&90.3&-&43.2&-&45.6&44.6& {53.0} &{60.5}\\
\multicolumn{2}{c|}{TransDA-B}  &88.9&46.1&85.4&11.6&1.7&50.7&61.3&38.8&88.4&-&93.7&74.4&30.0&89.6&-&49.4&-&49.3&49.2&56.8&65.0\\

\bottomrule  
\end{tabular}%
}
\vspace{-.5em}
\caption {``Best" results on Cityscapes test.}
\label{tab:lim}%
\end{table*}%

\subsection{Results on the Official Cityscapes Benchmark Server}

In addition, the current benchmarks of domain adaptive semantic segmentation also have some inherent problems, as discussed in \cite{araslanov2021self}'s supplement material. 


In fact, Cityscapes does not provide ground truth labeling for the test set, it instead provides a labeled validation set. 
The previous methods all verify the final model on the validation set, because the benchmarking process on the validation set does not have clear and strict standards, which is in discord with the established best practice on Cityscapes test set\cite{cordts2016cityscapes}, in particular.
The new practice evaluation is proposed and the results on Cityscapes test set are reported in \cite{araslanov2021self} supplement material, which encourages researchers to report the results on Cityscapes test set. 
The holdout test set for testing the final segmentation accuracy after adaptation becomes Cityscapes test, with the results obtained via submitting the predicted segmentation masks to the official Cityscapes benchmark server\footnote{\href{https://www.cityscapes-dataset.com}{https://www.cityscapes-dataset.com}}.
Owing to the regulated access to the test set, they believe this setting to offer more transparency and fairness to the benchmarking process.


However, Cityscapes Benchmark websites has a restricted access to the test annotation (\eg, limited number of submissions per time window and user), which limits researchers to consult the test set for verifying a number of model with different random seed. 
For this reason, we recommend that, for a fair comparison, use the validation set to select the best model 
and submit it for testing. 

The official Cityscapes benchmark server provides anonymous link to our results on Cityscapes test set for referencing in blind paper submissions. Table~\ref{tab:lim} shows the ``Best"  results on Cityscapes test for GTA5$\rightarrow$Cityscapes (TransDA-S\footnote{\href{https://www.cityscapes-dataset.com/anonymous-results/?id=b6ea23f38aa214510e3b69e2780af1e02d44f9d3c741e05842ac0756fae5dc4e}{https://www.cityscapes-dataset.com/anonymous-results/?id=\\b6ea23f38aa214510e3b69e2780af1e02d44f9d3c741e05842ac0756fae5dc4e}} and TransDA-B\footnote{\href{https://www.cityscapes-dataset.com/anonymous-results/?id=8dadcafb1549b277f5477246bde011c1ede443a84eecbf64ea261dfcbeec4d4f}{https://www.cityscapes-dataset.com/anonymous-results/?id=\\8dadcafb1549b277f5477246bde011c1ede443a84eecbf64ea261dfcbeec4d4f}}) and SYNTHIA$\rightarrow$Cityscapes (TransDA-S\footnote{\href{https://www.cityscapes-dataset.com/anonymous-results/?id=a805e7844fc51c7c7333e28a05529cbd19d23db058b5eff438819edfb623813b}{https://www.cityscapes-dataset.com/anonymous-results/?id=\\a805e7844fc51c7c7333e28a05529cbd19d23db058b5eff438819edfb623813b}} and TransDA-B\footnote{\href{https://www.cityscapes-dataset.com/anonymous-results/?id=102ba9c29c1fe0aa94c7ebd824b4cc80c6573ee80fe57db0cbe32222b1330e0f}{https://www.cityscapes-dataset.com/anonymous-results/?id=\\102ba9c29c1fe0aa94c7ebd824b4cc80c6573ee80fe57db0cbe32222b1330e0f}}).

\subsection{Datasets}
In this section, we provide more details about the datasets used in our experiments.

\paragraph{Cityscapes~\cite{cordts2016cityscapes}} contains 2975 real images in the training set, and the original image size is $2048 \times 1024$ pixels resolution. 
We utilize three data augmentation techniques to enhance the training stability, including random horizontal flipping, random re-scaling $1024 \times 512$ pixels within ratio range $[0.5, 2.0]$,  color jittering with brightness, contrast, saturation, and hue.

The two domain adaptive semantic segmentation tasks are GTA5$\rightarrow$Cityscapes and SYNTHIA$\rightarrow$Cityscapes. 
We use only the semantic classes shared with the simulation dataset for training and testing, and merge other categories as an ``ignore" class.
During the inference, the ``ignore" class area will be skipped. 
Following the previous protocol \cite{li2019bidirectional,wang2020classes,guo2021metacorrection,araslanov2021self,ma2021coarse,zhang2021prototypical}, we use 2975 images in the training set as the unlabeled target domain training set and evaluate the proposed model on 500 images in the validation set due to that its test set does not provide ground truth labeling.

\paragraph{GTA5~\cite{richter2016playing}} contains 24966 simulation images, and the original image size is $1914 \times 1052$ pixels resolution. 
We utilize three data augmentation techniques to enhance the training stability, including random horizontal flipping, random re-scaling $1280 \times 720$ pixels within ratio range $[0.5, 2.0]$,  color jittering with brightness, contrast, saturation, and hue.
Moreover, we employ only the 19 semantic classes shared with the real-world city street Cityscapes dataset.
Similarly, the other classes are all labelled as ``ignore'', and skipped during the inference.

\paragraph{Synthia-Rand-Cityscapes~\cite{ros2016synthia}} contains 9400 simulation images, the original image size is $1280 \times 760$ pixels resolution.
We utilize three data augmentation techniques to enhance the training stability, including random horizontal flipping, random re-scaling $1280 \times 760$ pixels within ratio range $[0.5, 2.0]$,  color jittering with brightness, contrast, saturation, and hue.
Similarly, only 16 semantic classes shared with the real-world city street Cityscapes dataset are used for training. 
Similarly, the other classes are all labelled as ``ignore'', and skipped during the inference.
However, this dataset usually selects the following two evaluation settings: performed on 16 classes or a subset of 13 classes. 
Here we follow the protocol in~\cite{pan2020unsupervised, wang2020classes, ma2021coarse} to train the model on the whole set and test it on both settings.

\subsection{Further Implementation Details}

\paragraph{Networks and Optimizers.} 
As for the feature extractor $\mathbf{F}$, we initially load the Swin-S model pre-trained on ImageNet-1K for TransDA-S, which has the similar model size and computation complexity to ResNet-101, and load the Swin-B model pre-trained on ImageNet-22K with the input size of 224 $\times$ 224 for TransDA-B, which is similar to ViT-B/DeiT-B in size.
The classifier $\mathbf{C}$ is implemented by UPerNet\cite{xiao2018unified} with deep supervision from FCN~\cite{long2015fully}. 
Following~\cite{liu2021Swin}, we use the AdamW~\cite{loshchilov2018decoupled} optimizer for the training of $\mathbf{F}$ as well as the classifier $\mathbf{C}$. 
We set the initial learning rate as $6 \times 10^{-5}$, weight decay as $0.01$, and employ a scheduler with linear learning rate decay along with a linear warm-up over $1500$ iterations. 
As for both the discriminator $\mathbf{D}$ and similarity network $\mathbf{S}$, we follow the settings from FADA~\cite{wang2020classes} and adopt a simple structure network consisting of 3 convolution layers. 
The Adam~\cite{kingma2014adam} optimizer is used for the training of  $\mathbf{D}$ and $\mathbf{S}$, where the initial learning rate is $1 \times 10^{-4}$, and the scheduler uses the `poly' learning rate decay with power $0.9$.

\paragraph{Reported Metrics.} For each experiments, we run each variant over five times with different random seeds unless otherwise stated. We use the single-scale test at the inference stage. 
Follow the common practice of previous works, we report the best mIoU for a fair comparison. Meanwhile, we also report the mean mIoU with standard deviations to demonstrate the robustness of TransDA.  
Since there are no ground-truth labels available for target domain in domain adaptive semantic segmentation, the optimal model cannot be picked out in practice.
Hence, we strongly prefer to show the robustness of our method by \color{red}\textbf{the averaged scores and standard deviations}.

\color{black}
\paragraph{Training Schedule.} 
Our models are trained on 4 Tesla V100 GPUs with 2 images per GPU per iteration, and the training process consists of one round of warm-up phase and three rounds of train phase, where each round lasts over 10k iterations.


\subsection{Broader Impact}

Our research can help reduce burden of collecting large-scale supervised data in many real-world applications of semantic segmentation by transferring knowledge from models trained on large labeled datasets to specific unlabeled datasets, especially sim2real scenario.  
The positive impact that our work could have on society is to make technology more accessible for institutions and individuals that do not have rich resources for annotating newly collected datasets. 
Besides, in this work, we put forward suggestions for the benchmarking of domain adaptive semantic segmentation, such as providing averaged score and standard deviation, and finally submitting the official Cityscapes benchmark score to standardize and promote this community.
We hope our research can also facilitate the thinking about the peculiarity of ViTs, and how it is potentially applicable to refresh the conventional transfer framework in the research fields.

\end{appendix}

\end{document}